\documentclass[3p]{elsarticle}

\usepackage{CJKutf8}
\usepackage{hyperref}
\usepackage[normalem]{ulem}
\usepackage{graphics}
\usepackage{amsmath,bm}
\usepackage{amssymb}
\usepackage{graphicx}
\usepackage{bm}
\usepackage{amsfonts}
\usepackage{color}
\usepackage{epstopdf}
\usepackage{booktabs}
\usepackage{array}
\usepackage{hyperref}
\usepackage{graphics}
\usepackage{epsfig}
\usepackage{bm}
\usepackage{epic,eepic}
\usepackage{epsfig}
\usepackage{pifont}

\usepackage{subfigure}

\usepackage{xspace}
\usepackage{graphicx,color}
\usepackage{amsmath,amssymb,amsfonts,mathrsfs}
\usepackage{url}
\usepackage[normalem]{ulem}
\usepackage{booktabs}
\usepackage{float}

\usepackage{tikz}
\usepackage{amsmath}
\usetikzlibrary{shapes, arrows.meta, positioning, calc, fit}

\usepackage[toc,page]{appendix}
\usepackage[capitalize]{cleveref}
\usepackage{mathtools}
\usepackage[normalem]{ulem}
\usepackage{gensymb}
\usepackage{algorithm}
\usepackage{algorithmic}

\newtheorem{thm}{Theorem}

\numberwithin{thm}{section} 
\newtheorem{rmk}[thm]{Remark}

\begin{document}
\begin{frontmatter}
\title{Equivariant U-Shaped Neural Operators for the Cahn–Hilliard Phase-Field Model}

\author[1]{Xiao Xue}
\author[2]{Marco F.P. ten Eikelder}
\ead{marco.eikelder@tu-darmstadt.de}
\author[1]{Tianyue Yang}
\author[4]{Yiqing Li}
\author[5]{Kan He}
\author[3]{Shuo Wang}
\author[1,6,7]{Peter V. Coveney}%
\ead{p.v.coveney@ucl.ac.uk}

\address[1]{Centre for Computational Science, Department of Chemistry, University College London, London, UK}
\address[2]{Institute for Mechanics, Computational Mechanics Group, Technical University of Darmstadt, Germany}
\address[3]{Department of Physics, Eindhoven University of Technology, Eindhoven, The Netherlands}
\address[4]{Department of Mechanical Engineering, University College London, London, UK}
\address[5]{College of Civil Engineering and Mechanics, Lanzhou University, Lanzhou 730000, China}
\address[6]{Informatics Institute, University of Amsterdam, Amsterdam, The Netherlands}
\address[7]{Centre for Advanced Research Computing, University College London, UK}


\date{\today}

\begin{abstract}
Phase separation in binary mixtures, governed by the Cahn–Hilliard equation, plays a central role in interfacial dynamics across materials science and soft matter. While numerical solvers are accurate, they are often computationally expensive and lack flexibility across varying initial conditions and geometries. Neural operators provide a data-driven alternative by learning solution operators between function spaces, but current architectures often fail to capture multiscale behavior and neglect underlying physical symmetries. Here we show that an equivariant U-shaped neural operator (E-UNO) can learn the evolution of the phase-field variable from short histories of past dynamics, achieving accurate predictions across space and time. The model combines global spectral convolution with a multi-resolution U-shaped architecture and regulates translation equivariance to align with the underlying physics. E-UNO outperforms standard Fourier neural operator and U-shaped neural operator baselines, particularly on fine-scale and high-frequency structures. By encoding symmetry and scale hierarchy, the model generalizes better, requires less training data, and yields physically consistent dynamics. This establishes E-UNO as an efficient surrogate for complex phase-field systems.

\end{abstract}


\end{frontmatter}
\section{Introduction}

Phase-field modeling has emerged as a powerful framework for multiphase and multicomponent systems \cite{anderson1998diffuse,kim2012phase,wight2020solving}. It addresses both physical modeling and geometrical representation through a smooth scalar phase-field variable that simultaneously represents a physical quantity (e.g., concentration or composition) and captures geometrical and topological changes in the interface between phases. Rather than requiring explicit tracking, interfaces are described as diffuse layers of finite thickness—a concept dating back to van der Waals and further developed by Cahn and Hilliard \cite{vanderWaals1979,cahn1958free}. The prototypical phase-field model that describes spinodal decomposition is the Cahn–Hilliard model. This is a fourth-order nonlinear partial differential equation (PDE) that models phase separation and coarsening (Ostwald ripening) in binary mixtures \cite{cahn1958free}. In the context of multiphase and multicomponent systems, the Cahn-Hilliard model takes a central place in the Navier–Stokes Cahn–Hilliard/Navier-Stokes Korteweg type models, both incompressible \cite{gurtinmodel,lowengrub1998quasi,ten2023unified,abels2024mathematical,ten2025unified,knopf2025thermodynamically} and compressible \cite{lowengrub1998quasi,ten2025compressible}, and is widely used in computations \cite{kay2007efficient,minjeaud2013unconditionally,ten2024divergence}. Besides multiphase flow applications, the Cahn-Hilliard model finds applications in a wide range of fields including metallurgy~\cite{cahn1958free}, tumor growth \cite{garcke2020long,ebenbeck2019analysis}, image analysis \cite{bertozzi2007analysis}, and pattern formation in biology \cite{khain2008generalized,elson2010phase,berry2018physical}.

Given the complexity and computational cost associated with solving such nonlinear partial differential equations, there is growing interest in leveraging machine learning (ML) to accelerate or augment traditional modeling approaches. Machine learning models have achieved remarkable success across various domains, including large language models~\cite{brown2020language}, computer vision~\cite{voulodimos2018deep}, and control tasks~\cite{ding2014data}. However, applying ML to the prediction of nonlinear dynamical systems - particularly those governed by partial differential equations, such as fluid dynamics~\cite{brunton2020machine} remains a significant challenge. To improve generalization in these settings, physics-informed machine learning~\cite{karniadakis2021physics,cheng2025machine,wang2025quantum} approaches have been proposed. These methods incorporate physical laws or equations directly into the learning process, often through modified loss functions or architecture constraints, enabling applications in turbulent flows~\cite{fukami2019synthetic,xue2022synthetic}, near-wall modeling~\cite{yang2019predictive,xue2024physics}, and heat transfer problems. Nevertheless, embedding PDE constraints into the training objective can substantially increase computational costs, limiting scalability and practical deployment for real-world scenarios. Generative models, such as Generative Adversarial Networks (GANs), have also been explored for dynamical system prediction, but they often struggle with accuracy and stability~\cite{cheng2020data}. More recently, neural operators have been developed as data-driven models that approximate mappings between infinite-dimensional function spaces, trained on data produced by traditional numerical solvers for PDEs. Once trained, they can act as fast surrogates that accelerate PDE-based simulations, though their reliability remains problem-dependent and requires careful assessment~\cite{Li2020FourierNO}. Among them, Deep Operator Networks (DeepONet)~\cite{lu2021learning} introduced a branch–trunk architecture to approximate nonlinear operators directly from data, while the Fourier Neural Operator (FNO)~\cite{Li2020FourierNO} and its successors, such as the Laplace Neural Operator (LNO)~\cite{cao2024laplace}, provide mesh-independent frameworks capable of capturing complex dynamics with high efficiency and generalization. These approaches have shown impressive performance across a wide range of PDE-based dynamical systems.

Despite this progress, the application of machine learning within the phase-field modeling community remains in its early stages. Recent efforts have begun to bridge this gap: Shen et al.\cite{shen2019phase} employed machine learning to derive analytical expressions from high-throughput phase-field simulations; Feng et al.\cite{feng2021machine} integrated a surrogate model to accelerate crack growth simulations; Teichert et al.\cite{TEICHERT2019201} applied an Integrable Deep Neural Network (IDNN) for defect formation prediction; and Oommen and co-workers~\cite{oommen2024rethinking} proposed a promising way to hybrid numerical solvers with machine learning models. While promising, these studies highlight that ML-enhanced phase-field modeling still faces accuracy and scalability challenges. One promising direction to address these limitations is the use of equivariance representations, which enforce consistency of model outputs under symmetry transformations inherent to physical systems. Equivariant neural networks have shown strong performance in fields such as computer vision~\cite{cohen2016group}, molecular modeling and drug discovery~\cite{thomas2018tensor, schutt2021equivariant}, where group symmetries (e.g., rotation, reflection) are naturally present. However, their application in physics-informed machine learning remains limited, particularly in complex PDE-based systems such as phase-field models. 
 
To address the current challenges in applying machine learning to phase-field modeling, we propose an Equivariant U-shaped Neural Operator (E-UNO) framework that incorporates the discrete symmetries of the governing equations. By integrating dihedral group-equivariance into the neural operator framework, our approach aligns the model architecture with the inherent symmetries of the governing physics, enhancing accuracy, generalization, and data efficiency. The U-shaped structure further combines hierarchical encoding with spectral representations, enabling super-resolution capabilities essential for capturing fine-scale microstructural features. We conduct a comprehensive evaluation of the proposed E-UNO model in the context of microstructural evolution prediction, benchmarking its performance against both the FNO and the standard UNO. The results show that our E-UNO systematically outperforms both models. In addition, we propose the incorporation of gradient-based constraints into the learning framework to better enforce the underlying physical principles and improve the interpretability of the model. We also explore multiple prediction strategies to assess robustness and provide a comprehensive analysis of model behavior in different settings. Through this investigation, our research aims to advance the integration of neural operators with phase-field methods, contributing to the development of efficient, accurate, and physics-aware surrogate models for scientific applications.

This paper is organized as follows. In~\cref{sec:cahn-hilliard}, we present the theoretical foundations of the Cahn-Hilliard model. In~\cref{sec: equivariant NO framework} we provide the framework of the equivariant U-shaped Fourier neural operator framework. The numerical results on the prediction accuracy and model performance are discussed in~\cref{sec:results}. Finally, conclusions are drawn in~\cref{sec:conclusion}.

\section{The Cahn-Hilliard model}\label{sec:cahn-hilliard}
The Cahn-Hilliard (CH) model is classically derived as the gradient-flow of a free energy functional \cite{novick2008cahn}, and also naturally emerges through continuum mixture theory in combination with the Colemann–Noll procedure \cite{ten2023unified}. It governs the evolution of an order parameter $\Phi: \Omega \times \mathcal{T} \rightarrow \mathbb{R}$, where $\Omega \subset \mathbb{R}^d$ is a simply connected domain with dimension $d=2,3$, and $\mathcal{T}=(0,T)$ is the time domain with end time $T>0$. The order parameter $\Phi=\Phi(\mathbf{x},t)$ describes the composition of the binary mixture under consideration. Its physically-admissible range is $\Phi \in [-1,1]$, where $\Phi = \pm 1$ correspond to the two pure components, and $-1<\Phi<1$ represents a diffuse interface layer with a mixture of the two components.
\begin{rmk}[Range order parameter]
To describe the composition of a binary mixture there are various physical quantities that one can adopt. Typical choices are concentrations and volume fractions. These quantities range between $0$ and $1$. The current order parameter $\Phi$ describes the difference between such a quantity of the first and second component, its physically-admissible range is $\Phi \in [-1,1]$.
\end{rmk}

We consider the basic formulation of the (isothermal) CH equation with the gradient-flow structure:
\begin{subequations}\label{eq:Cahn}
\begin{align}
    \frac{\partial \Phi}{\partial t}  =&~ {\rm div}\left( \gamma \nabla \mu \right), \label{eq:PDE}\\
    \mathbf{n}\cdot\nabla \Phi =&~ 0, \label{eq:BC1}\\\mathbf{n}\cdot\nabla \mu =&~ 0, \label{eq:BC2}
\end{align}
\end{subequations}
where \eqref{eq:BC1} and \eqref{eq:BC2} are homogeneous Neumann-type boundary conditions with $\mathbf{n}$ the unit outward normal. In \eqref{eq:PDE} the quantity $\gamma=\gamma(\Phi) \geq 0$ denotes the mobility that determines the time scale of the phase separation and coarsening processes. Furthermore, $\mu$ is the chemical potential defined as the variational derivative of the Helmholtz free energy functional $F$:
\begin{align}\label{eq:Chem Pot}    
    \mu = \frac{\delta F}{\delta \Phi}.
\end{align}
The conservation of the order parameter follows from the integration of \eqref{eq:PDE}:
\begin{align}
    \int_\Omega \Phi(\mathbf{x},t) ~{\rm d}v = \int_\Omega \Phi(\mathbf{x},0) ~{\rm d}v,
\end{align}
for $(\mathbf{x},t)\in \Omega_T$, where $\Omega_T = \Omega \times \mathcal{T}$ and where we have utilized Gauss' theorem and \eqref{eq:BC2}, and where ${\rm d}v$ is the volume element. Additionally, the system \eqref{eq:Cahn} is equipped with the energy dissipation:
\begin{align}\label{eq:Energy}    
    \dfrac{{\rm d}}{{\rm d}t} F = \int_\Omega \frac{\delta F}{\delta \Phi} \frac{\partial \Phi}{\partial t} ~{\rm d}v = \int_\Omega \mu {\rm div}\left( \gamma \nabla \mu \right) ~{\rm d}v = - \int_\Omega \gamma \|\nabla \mu\|^2 ~{\rm d}v \leq 0, 
\end{align}
where we have invoked \eqref{eq:BC2}. We emphasize that \eqref{eq:Energy} is contingent on the positivity of the mobility quantity $\gamma \geq 0$.

\begin{figure}[h]
    \centering
    \includegraphics[width=0.5\linewidth]{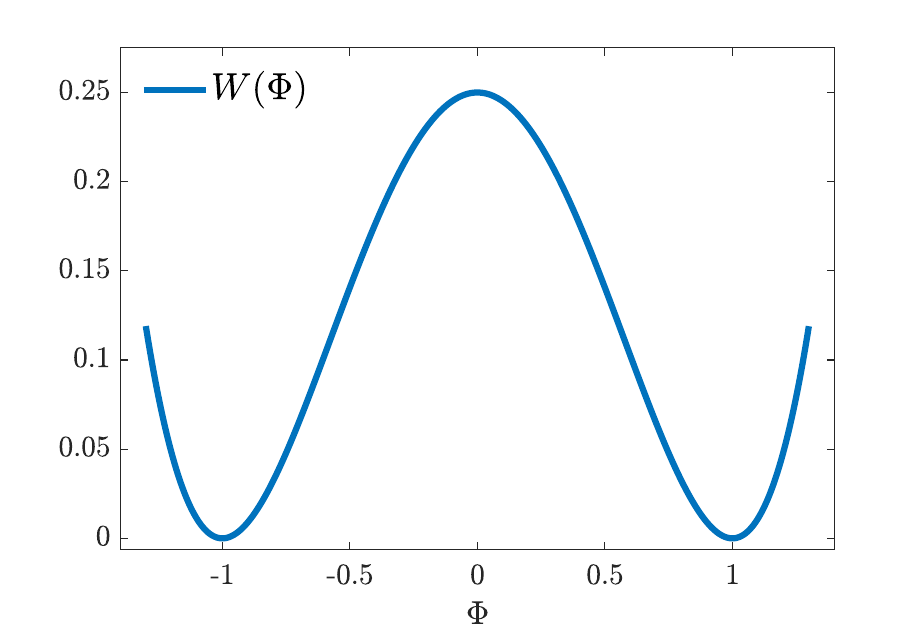}
    \caption{Ginzburg-Landau free energy potential $W=W(\Phi)$. The free energy potential has a double well structure with minima at the pure phases represented by $\Phi = \pm 1$.}
    \label{fig: double well}
\end{figure}

We consider the Cahn-Hilliard system \eqref{eq:Cahn} for the polynomial Ginzburg-Landau Helmholtz free energy potential:
\begin{subequations}\label{eq: free energy}
\begin{align}
    F =&~ \int_\Omega f~{\rm d}v,\\
    f =&~\frac{\lambda}{\epsilon}W(\Phi) + \frac{1}{2}\lambda \epsilon \|\nabla  \Phi \|^2,\\
    W(\Phi) =&~ \frac{1}{4}(\Phi^2-1)^2,
\end{align}    
\end{subequations}
where the quantity $\epsilon > 0$ is an interface thickness parameter and $\lambda > 0$ is a surface energy density parameter. The free energy density $f$ contains the double-well potential $W=W(\Phi)$ (visualized in FIG. \ref{fig: double well}) that favors phase separation, and the gradient term that favors coarsening. 

The corresponding chemical potential takes the form:
\begin{align}\label{eq: chem GL}
   \mu =  \lambda \left(\frac{\Phi(\Phi^2-1)}{\epsilon}-\epsilon\Delta^2\Phi\right).
\end{align}

\begin{figure}[h!]
\centering
\subfigure[Equilibrium profile $\Phi^{\rm eq}=\Phi^{\rm eq}(s)$]{\includegraphics[width=0.49\textwidth]{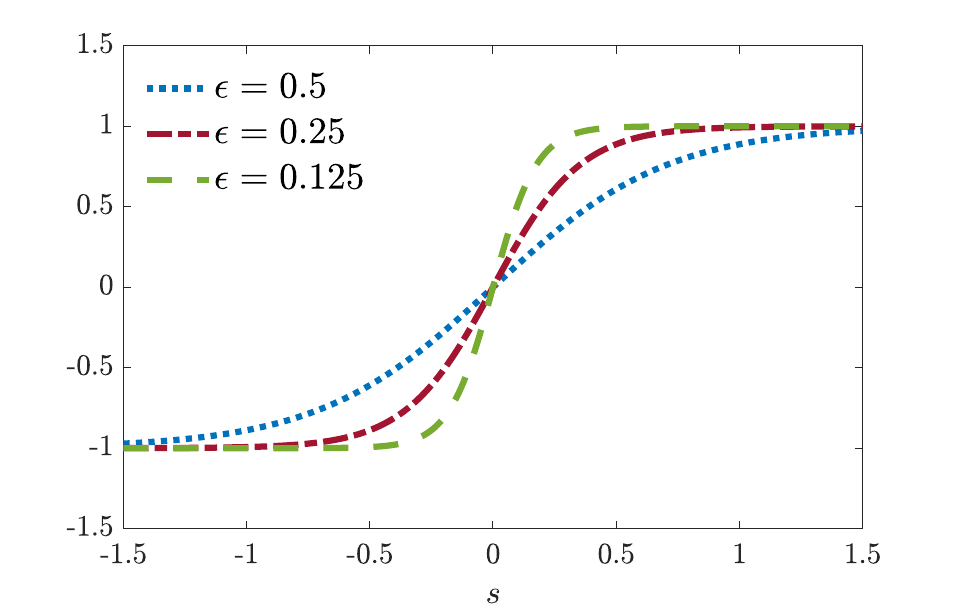}}\label{fig:equi prof}
\subfigure[Equilibrium free energy density $f^{\rm eq}=f^{\rm eq}(s)$] {\includegraphics[width=0.49\textwidth]{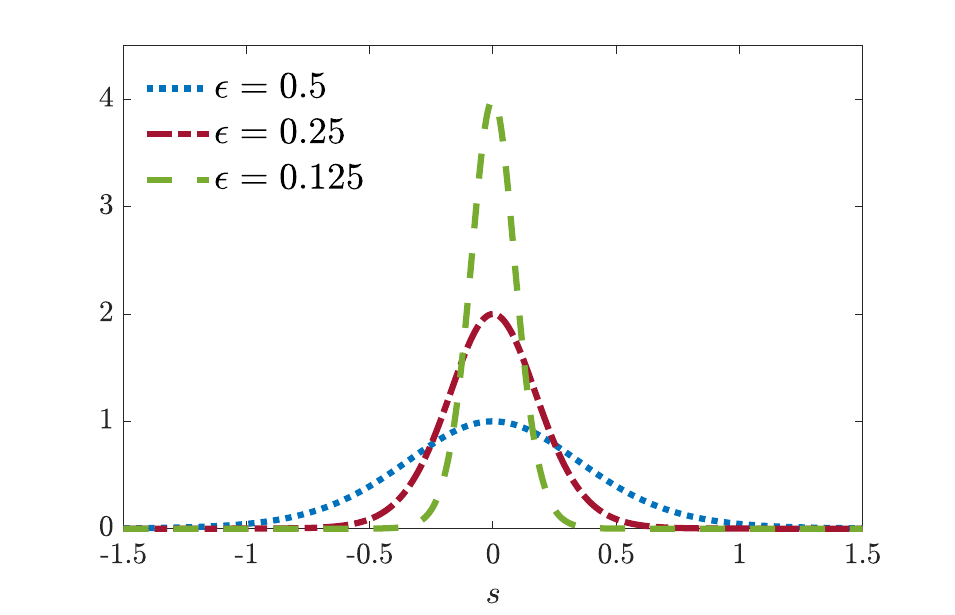}}\label{fig:equi energy}
\caption{In one spatial dimension (coordinate $s$) the Cahn-Hilliard equation has an equilbrium solution with an analytical expression. Panel (a) shows the equilibrium profile $\Phi^{\rm eq}=\Phi^{\rm eq}(s)$, and panel (b) shows the corresponding free energy density. In both panels the interface width parameter is varied.} \label{fig:equilibrium}
\end{figure}

The equilibrium solution of the Cahn-Hilliard system \eqref{eq:Cahn} is governed by a constant chemical potential $\mu = {\rm const}$, where ${\rm const}$ does not depend on $(\mathbf{x},t) \in \Omega \times \mathcal{T}$. Inserting the chemical potential \eqref{eq: chem GL}, the condition becomes:
\begin{align}\label{eq: equilibrium}
   \lambda \left(\frac{\Phi^{\rm eq}\left((\Phi^{\rm eq})^2-1\right)}{\epsilon}-\epsilon \Delta^2\Phi^{\rm eq}\right) = {\rm const}.
\end{align}
In one-dimension, one may verify that the equilibrium profile $\Phi=\Phi^{\rm eq}$:
\begin{align}
    \Phi^{\rm eq} = \tanh\left( \dfrac{s}{\epsilon \sqrt{2}} \right).
\end{align}
satisfies the condition \eqref{eq: equilibrium} with ${\rm const} = 0$, where $s$ is the spatial coordinate centered at the interface $\Phi=0$. In this case the equilibrium free energy $F=F^{\rm eq}$, and equilibrium free density density $f=f^{\rm eq}$ take respectively the forms:
\begin{subequations}
    \begin{align}
        f^{\rm eq}  =&~ \frac{\lambda}{2\epsilon} {\rm sech}^4\left(\frac{s}{\epsilon\sqrt{2}}\right),\\
        F^{\rm eq} =& \lambda \frac{2\sqrt{2}}{3}.
    \end{align}
\end{subequations}
We visualize the equilibrium profile and free energy density in FIG~\ref{fig:equilibrium}. We observe that in the sharp-interface limit $\epsilon \rightarrow 0$ the interface profile converges towards a Heaviside function, and the equilibrium free energy density collapses onto the
interface ($s=0$).

\begin{rmk}[Mobility]
    There are several forms for the mobility quantity $\gamma$ that appear in the literature. For the sake of simplicity, the mobility is often taken as a constant (in space and time). Alternatively, it may depend on the mixture composition and vanishes in the single phase regime, i.e. it is degenerate. This implies that pure phases have no mobility and the right-hand side of \eqref{eq:Cahn} vanishes there. In some scenarios, thermodynamic considerations indicate that a degenerate mobility is a physically appropriate choice \cite{novick2008cahn}. In addition, the mobility may depend on $\epsilon$. In general, stability arguments show that the mobility should be a positive (greater or equal zero) quantity.
\end{rmk}

\section{Equivariant Neural Operator framework}\label{sec: equivariant NO framework}
In this section we discuss the application of the Neural operator framework to the Cahn-Hilliard equation.

\subsection{Neural operator framework}\label{subsec: NO framework}
We begin by introducing the Neural Operator (NO) framework, which aims to learn mappings (the NO) between infinite-dimensional function spaces\cite{kovachki2023neural}. This generalizes classical neural networks, which map between finite-dimensional vectors, to the setting where both the input and output are functions defined over a spatial domain. Learning the entire solution operator—rather than just a single solution—makes neural operators especially appealing for problems in scientific computing, where the solution operator itself plays a central role. This operator-centric perspective enables neural operators to generalize across varying inputs, boundary conditions, and even geometries, making them data-efficient and highly transferable. 

We consider the task of approximating a nonlinear operator that maps an input function, possibly defined over both space and time, to a function-valued output. Formally, let $\Omega \subset \mathbb{R}^d$ denote the spatial domain with coordinates $\mathbf{x} \in \Omega$ and $0<t<T$ the time with final time $T > 0$. Let $\mathcal{G}^{\dagger}$ be a (possibly nonlinear) solution operator acting between separable Banach spaces $\mathcal{A}$ and $\mathcal{U}$:
\begin{subequations}
  \begin{align}
    \mathcal{G}^{\dagger} : \mathcal{A} &\rightarrow \mathcal{U}, \\
    a &\mapsto u,
  \end{align}
\end{subequations}
where $a \in \mathcal{A}$ represents an input function (possibly time-dependent), and $u \in \mathcal{U}$ is the corresponding output function of interest. We use superscripts $\dagger$ to denote true quantities (operators/parameters). This setting encompasses a broad class of operator learning problems, including steady-state, time-dependent, and parameterized PDEs. 

To approximate the solution operator $\mathcal{G}^{\dagger}$, we first suppose that $a_j$ are independent and identically distributed drawn from probability measure $\mu$ in $\mathcal{A}$ and $u_j = \mathcal{G}^{\dagger}\left(a_j\right)$. The neural operator $\mathcal{G}_{\theta}$ is now constructed as a parametric map:
  \begin{align}\label{eq: neural operator}
    \mathcal{G}_\theta :&~ \mathcal{A} \mapsto \mathcal{U},
\end{align}
with parameter $\theta \in \mathbb{R}^p$. We aim to build $\mathcal{G}_\theta$ so that  $\mathcal{G}_{\theta^\dagger} \approx \mathcal{G}^{\dagger}$, more specifically $\mathcal{G}_\theta$ learns an approximation of $\mathcal{G}^{\dagger}$ by minimizing
\begin{equation}
    \theta^\dagger = {\arg\min}_\theta \mathbb{E}_{a \sim \mu}\left[\mathcal{L}\left(\mathcal{G}_\theta(a), \mathcal{G}^{\dagger}(a)\right)\right]
\end{equation}
where the cost function $\mathcal{L}:\mathcal{U}\times\mathcal{U}\rightarrow \mathbb{R}_+$ is suitable norm.

For the Cahn-Hilliard evolution equation with homogeneous/periodic boundary conditions, we aim to learn a surrogate model that predicts the evolution of solution $\Phi=\Phi(\mathbf{x},t)$ over a future time interval $t\in[\tau, \tau + \tau_{\rm out}]$, given its history over a prior time interval $t \in [\tau - \tau_{\rm in}, \tau]$. In this case, the solution operator is given by:
\begin{subequations}
  \begin{align}
    \mathcal{G}^{\dagger} : \mathcal{A} &\rightarrow \mathcal{U}, \\
    \mathcal{G}^{\dagger}\left(\Phi(\cdot,\tau)\big|_{\tau \in [t - t_{\text{in}},\, t]}\right) &= \Phi(\cdot,\tau)\big|_{\tau \in [t,\, t + t_{\text{out}}]},
  \end{align}
\end{subequations}
where $\mathcal{A} = L_2([t - t_{\text{in}},\, t]; H^2(\Omega))$ and $\mathcal{U} = L_2([t,\, t + t_{\text{out}}]; H^2(\Omega))$ are Bochner spaces of time-indexed spatial functions \cite{evans2022partial}. Hence, in the context of our evolution equation, we have $a_j=\Phi(\mathbf{x}_j,\tau_j)$ with $\mathbf{x}_j \in \Omega$, $\tau_j \in [t-t_{\rm in}]$ and $u_j = \mathcal{G}^{\dagger}(\Phi(\mathbf{x}_j,\tau_j))$. The neural operator is now constructed by finding the parameters $\theta=\theta^\dagger$ that minimize the distance between applying $\mathcal{G}_\theta$ on the space-time function samples $a_j=\Phi(\mathbf{x}_j,\tau_j)$ and the actual solutions:
\begin{align}
    \theta^\dagger = {\arg\min}_{\theta \in \mathbb{R}^p} \frac{1}{N} \sum_{j=1,...,N} \mathcal{L}(\mathcal{G}_\theta(\Phi(\mathbf{x}_j,\tau_j)),\mathcal{G}^\dagger(\Phi(\mathbf{x}_j,\tau_j))),
\end{align}
where $N$ is the number of samples.

The neural operator $\mathcal{G}_\theta$ is constructed as a composition of the maps:
\begin{align}
    \mathcal{G}_\theta = \mathcal{Q} \circ \mathcal{K}^{(S)}_\theta \circ \cdots \circ \mathcal{K}^{(1)}_\theta \circ \mathcal{P},
\end{align}
where the lifting operator $\mathcal{P}$ maps $a \in \mathcal{A}$ to a feature field $v_0: \Omega \to \mathbb{R}^{d_0}$, the kernel layers iteratively update $v_s: \Omega \to \mathbb{R}^{d_s}$ as $v_{s}=\mathcal{K}^s_\theta v_{s-1}$, $s=1,...,S$, and the projection $\mathcal{Q}$ maps $v_S: \Omega \to \mathbb{R}^{d_S}$ to $\mathcal{U}$. 

The lifting operator $\mathcal{P}$ increases the expressive capacity of the model by embedding the input function into a higher-dimensional latent feature space. This enables the network to encode complex local patterns and abstract representations that may be difficult to capture in the original input space. The sequence of kernel layers $\mathcal{K}_\theta^{(s)}$ then propagates and transforms these features through a hierarchy of nonlinear operations. Each kernel layer may incorporate both local and global interactions across the spatial domain, depending on the chosen architecture. Finally, the projection operator $\mathcal{Q}$ maps the learned latent representation to the target output space. It acts pointwise, reducing the feature dimensionality to match that of the desired output function. This structure enables the neural operator to approximate a wide class of nonlinear, possibly nonlocal, solution operators, while preserving resolution and mesh-independence.

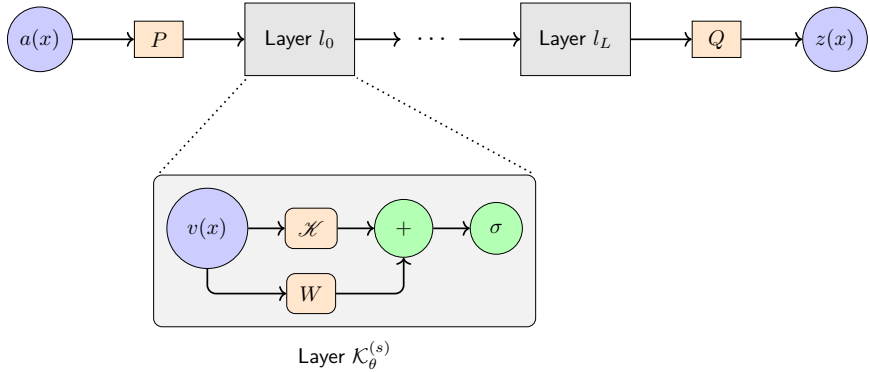
\begin{figure}[h]
\begin{center}
\resizebox{0.7\textwidth}{!}{%
    \begin{tikzpicture}[
    node distance=0.8cm and 1cm,
    every node/.style={font=\sffamily},
    round/.style={circle, draw, fill=blue!20, minimum size=8mm},
    box/.style={rectangle, draw, fill=orange!20, minimum width=8mm, minimum height=6mm},
    layer/.style={rectangle, draw, fill=gray!20, minimum height=1.2cm, minimum width=1.8cm, align=center},
    flow/.style={->, thick},
    dots/.style={minimum width=6mm, align=center, font=\large, draw=none, fill=none, inner sep=0pt}
]

\node[round] (a) {$a(x)$};
\node[box, right=of a] (P) {$P$};
\node[layer, right=of P] (L0) {Layer $l_0$};
\node[dots, right=of L0] (L1) { $\cdots$ };
\node[layer, right=of L1] (LL) {Layer $l_L$};
\node[box, right=of LL] (Q) {$Q$};
\node[round, right=of Q] (z) {$z(x)$};

\foreach \from/\to in {a/P, P/L0, LL/Q, Q/z}
    \draw[flow] (\from) -- (\to);
\draw[flow] (L0.east) -- ++(0.75,0);
\draw[flow] (L1.east) -- ++(1,0);

\coordinate (L0-bl) at ([xshift=-2.3cm,yshift=-1.6cm]L0.south);
\coordinate (L0-br) at ([xshift= 3.8cm,yshift=-1.6cm]L0.south);

\begin{scope}[xshift=5cm, yshift=-3.5cm]
\node[draw, fill=gray!10, inner sep=6pt, rounded corners] (panelb) {
    \begin{tikzpicture}[node distance=0.4cm and 0.6cm]
        \node[round] (vin) {$v(x)$};

        \node[box, right=of vin] (Fblock) {$\mathscr{K}$};
        \node[box, below=of Fblock] (W) {$W$};
        \node[circle, draw, fill=green!30, right=of Fblock] (add) {$+$};
        \node[circle, draw, fill=green!30, right=of add] (sigma) {$\sigma$};
        \draw[flow] (vin) -- (Fblock.west);     
        \draw[flow] (Fblock.east) -- (add);     
        \draw[flow] (vin) |- (W);               
        \draw[flow] (W) -| (add);               
        \draw[flow] (add) -- (sigma);               
    \end{tikzpicture}
};
\node[below=0.1cm of panelb] {Layer $\mathcal{K}_\theta^{(s)}$};
\coordinate (panelb-tl) at ([xshift=0.2cm,yshift=1.6cm]panelb.north);
\coordinate (panelb-tr) at ([xshift= -1.6cm,yshift=1.6cm]panelb.north);
\end{scope}

\draw[dotted, thick] (L0-bl) -- (panelb-tr);
\draw[dotted, thick] (L0-br) -- (panelb-tl);

\end{tikzpicture}
    }
\caption{Schematic illustration of the neural operator architecture. The mapping from input functions $a(x)$ to output functions $z(x)$. Each layer combines a linear transformation $W$ with a nonlocal integral operator $\mathscr{K}$ followed by a nonlinearity $\sigma$.}
\end{center}
\label{fig: NO architecture}
\end{figure}

In general, each kernel layer $\mathcal{K}_\theta^{(s)}$ can be interpreted as a nonlinear operator that transforms the feature field by aggregating information across the spatial domain. A common design is to represent this transformation as an integral operator:
\begin{subequations}
\begin{align}
    v_s(\mathbf{x}) =&~ \sigma\left( W v_{s-1}(\mathbf{x}) + \mathscr{K}_\theta^{(s)}v_{s-1}(\mathbf{x}) \right),\\    
    \mathscr{K}_\theta^{(s)}v_{s-1}(\mathbf{x}) =&~ \int_{\Omega} \kappa_\theta^{(s)}(\mathbf{x}, \mathbf{y}) v_{s-1}(\mathbf{y}) \, \mathrm{d}\mathbf{y},
\end{align}    
\end{subequations}
where $\kappa_\theta^{(s)}: \Omega \times \Omega \to \mathbb{R}^{d_s \times d_{s-1}}$ is a learned kernel function that captures interactions between spatial points, $W \in \mathbb{R}^{d_s \times d_{s-1}}$ is a pointwise linear transformation, and $\sigma$ is a nonlinear activation. This structure allows each layer to model both local and nonlocal dependencies, which are essential in many PDE-driven problems. We sketch the architecture in Figure \ref{fig: NO architecture}.

Different neural operator architectures correspond to different choices for how the kernel $\kappa_\theta$ is parameterized and applied. The specific realization of these kernel layers is what distinguishes, for example, integral operator networks, attention-based operators, and spectral neural operators. In the next section, we present one such realization: the Fourier Neural Operator.

\subsection{Fourier Neural Operators}
The Fourier Neural Operator (FNO)~\cite{li2020fourier} is a specialization of the general neural operator framework introduced in~\cref{subsec: NO framework}, where the kernel network $\mathcal{K}_\theta$ is implemented via spectral convolution layers. This construction leverages the Fourier transform to efficiently encode global spatial interactions, making it particularly well-suited for approximating solution operators of PDEs that exhibit long-range dependencies and smooth fields, as commonly found in phase-field models.

The core idea of the FNO is to leverage the convolution theorem: convolution in the spatial domain corresponds to multiplication in the Fourier domain. Instead of explicitly learning a spatial kernel $\kappa_\theta(x,y)$ as in classical integral operators, the FNO applies a parameterized, mode-wise transformation to the Fourier coefficients of the feature field

We now consider a specific realization of the neural operator architecture in which each kernel layer is modeled as a translation-invariant integral operator. More precisely, we assume that the kernel depends only on the relative distance between spatial points, i.e.,
\begin{align}
    \kappa_\theta^{(s)}(\mathbf{x}, \mathbf{y}) = \kappa_\theta^{(s)}(\mathbf{x} - \mathbf{y}),
\end{align}
where, with a slight abuse of notation, we write the kernel as a function of the offset $\mathbf{x} - \mathbf{y}$. Under this assumption, the integral operator becomes a convolution:
\begin{align}
    v_s(\mathbf{x}) = \sigma\left( W v_{s-1}(\mathbf{x}) + \left( \kappa_\theta^{(s)} * v_{s-1} \right)(\mathbf{x}) \right),
\end{align}
where $*$ denotes the convolution over the spatial domain $\Omega$. Next, to compute this convolution efficiently, we apply the discrete Fourier transform (DFT) to the convolution term. By the convolution theorem,
\begin{align}
    \mathcal{F}\left[ \kappa_\theta^{(s)} * v_{s-1} \right](\mathbf{k}) = \mathcal{F}\left[ \kappa_\theta^{(s)} \right](\mathbf{k}) \cdot \mathcal{F}\left[ v_{s-1} \right](\mathbf{k}).
\end{align}
where $\mathbf{k}$ is the multi-index of frequency modes. The FNO approximates this frequency-wise multiplication by learning the Fourier representation directly. That is, the operator $\tilde{\mathscr{K}}_\theta^{(s)}$ approximates $\mathscr{K}_\theta^{(s)}$ and is defined via:
\begin{align}
    \tilde{\mathscr{K}}_\theta^{(s)} v_{s-1} = \mathcal{F}^{-1}[\mathcal{R}_\theta^{(s)}(\mathbf{k}) \cdot \mathcal{F}\left[ v_{s-1} \right](\mathbf{k})],
\end{align}
where $\mathcal{R}_\theta^{(s)}(\mathbf{k}) \in \mathbb{C}^{d_s \times d_{s-1}}$ is a learned matrix for each mode $\mathbf{k}$ with $\|\mathbf{k}\|_\infty \leq k_{\max}$, and zero otherwise, for some threshold $k_{\max}>0$. Hence, all higher frequency modes are zeroed out and the corresponding spectral multiplication can be written componentwise as
\begin{equation}
    \left(\mathcal{R}_\theta^{(s)} \cdot \mathcal{F}(v_{s-1})\right)_{\mathbf{k}, i}
    = \sum_{j=1}^{d_{s-1}} \mathcal{R}_{\mathbf{k}, i, j}^{(s)} \, \mathcal{F}(v_{s-1})_{\mathbf{k}, j}, \quad \forall \|\mathbf{k}\|_\infty \leq k_{\max},\ i = 1, \dots, d_s.
\end{equation}

\begin{figure}[h]
\begin{center}
\resizebox{0.7\textwidth}{!}{%
    \begin{tikzpicture}[
    node distance=0.8cm and 1cm,
    every node/.style={font=\sffamily},
    round/.style={circle, draw, fill=blue!20, minimum size=8mm},
    box/.style={rectangle, draw, fill=orange!20, minimum width=8mm, minimum height=6mm},
    layer/.style={rectangle, draw, fill=gray!20, minimum height=1.2cm, minimum width=1.8cm, align=center},
    flow/.style={->, thick},
    dots/.style={minimum width=6mm, align=center, font=\large, draw=none, fill=none, inner sep=0pt}
]

\node[round] (a) {$a(x)$};
\node[box, right=of a] (P) {$P$};
\node[layer, right=of P] (L0) {Fourier\\layer $l_0$};
\node[dots, right=of L0] (L1) { $\cdots$ };
\node[layer, right=of L1] (LL) {Fourier\\layer $l_L$};
\node[box, right=of LL] (Q) {$Q$};
\node[round, right=of Q] (z) {$z(x)$};

\foreach \from/\to in {a/P, P/L0, LL/Q, Q/z}
    \draw[flow] (\from) -- (\to);
\draw[flow] (L0.east) -- ++(0.75,0);
\draw[flow] (L1.east) -- ++(1,0);

\coordinate (L0-bl) at ([xshift=-3.8cm,yshift=-1.5cm]L0.south);
\coordinate (L0-br) at ([xshift= 5.3cm,yshift=-1.5cm]L0.south);

\begin{scope}[xshift=5cm, yshift=-3.5cm]
\node[draw, fill=gray!10, inner sep=6pt, rounded corners] (panelb) {
    \begin{tikzpicture}[node distance=0.4cm and 0.6cm]
        \node[round] (vin) {$v(x)$};
        \node[draw, rectangle, minimum width=2cm, minimum height=1cm, right=of vin, fill=yellow!20] (Fblock) {
            \begin{tikzpicture}[node distance=0.2cm and 0.4cm]
                \node[box] (F) {$\mathcal{F}$};
                \node[box, right=of F] (R) {$R$};
                \node[box, right=of R] (Fi) {$\mathcal{F}^{-1}$};
            \end{tikzpicture}
        };
        \node[box, below=of Fblock] (W) {$W$};
        \node[circle, draw, fill=green!30, right=of Fblock] (add) {$+$};
        \node[circle, draw, fill=green!30, right=of add] (sigma) {$\sigma$};
        \draw[flow] (vin) -- (Fblock.west);     
        \draw[flow] (Fblock.east) -- (add);     
        \draw[flow] (vin) |- (W);               
        \draw[flow] (W) -| (add);               
        \draw[flow] (add) -- (sigma);               
    \end{tikzpicture}
};
\node[below=0.1cm of panelb] {Fourier layer};
\coordinate (panelb-tl) at ([xshift=0.1cm,yshift=1.5cm]panelb.north);
\coordinate (panelb-tr) at ([xshift= -1.6cm,yshift=1.5cm]panelb.north);
\end{scope}

\draw[dotted, thick] (L0-bl) -- (panelb-tr);
\draw[dotted, thick] (L0-br) -- (panelb-tl);

\end{tikzpicture}
    }
\caption{Schematic of the Fourier Neural Operator (FNO) architecture. The mapping from input functions $a(x)$ to output functions $z(x)$ is parameterized through stacked Fourier layers $\{l_0, \dots, l_L\}$. Each layer consists of a linear transformation $W$ together with a spectral convolution obtained by   applying a Fourier transform $\mathcal{F}$, multiplying by learned spectral weights ($R$), and applying the inverse transform $\mathcal{F}^{-1}$, followed by nonlinearity $\sigma$. This enables learning of nonlocal operators directly in frequency space.}
\end{center}
\label{fig: FNO architecture}
\end{figure}
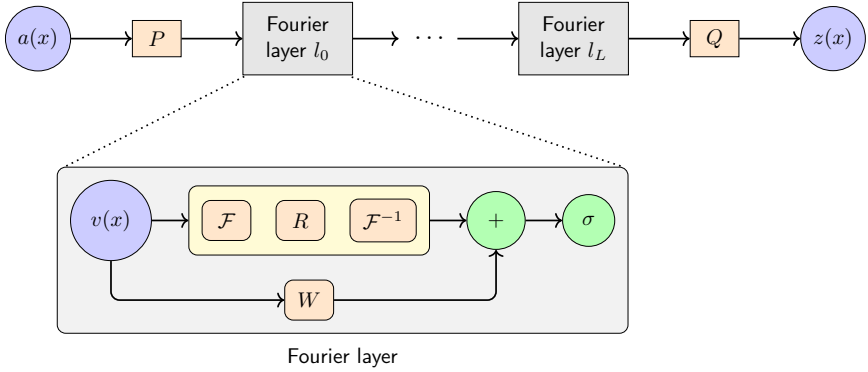

The action of the approximated kernel operator is then computed as follows. First, the discrete Fourier transform of the input feature field $v_{s-1}$ is computed, yielding $\mathcal{F}[v_{s-1}](\mathbf{k})$. Next, each mode is multiplied by a learned weight matrix, resulting in $\mathcal{F}[\tilde{\mathscr{K}}_\theta^{(s)} v_{s-1}](\mathbf{k}) = \mathcal{R}_\theta^{(s)}(\mathbf{k}) \cdot \mathcal{F}[v_{s-1}](\mathbf{k})$, where the transformation $\mathcal{R}_\theta^{(s)}(\mathbf{k})$ is learned and truncated to a finite number of modes with $\|\mathbf{k}\|_\infty \leq K$. Finally, the inverse Fourier transform is applied to return to the spatial domain, yielding the approximation $\tilde{\mathscr{K}}_\theta^{(s)} v_{s-1}(\mathbf{x})$. This spectral path, illustrated in Figure \ref{fig: FNO architecture}, enriches the representational capacity of the model, especially in the low- to mid-frequency bands. By parameterizing the kernel action directly in Fourier space, the FNO avoids spatial convolution entirely, while preserving global coupling and mesh-independence.

\subsection{Equivariant UNO Framework}\label{sec:E-UNO}

To incorporate the physical symmetries of the phase-field process into the learning framework, we introduce an equivariance loss based on the dihedral group $D_4$, which consists of the eight symmetries of a square, including rotations and reflections. The loss enforces that the neural operator prediction commutes with the action of the group. Specifically, for a neural operator $\mathcal{G}$ that evolves the phase field $\phi_t$ to $\phi_{t+1}$, we define the equivariance loss as
\begin{align}
\mathcal{L}_{\text{eq}} = \sum_{g \in D_4} \sum_{\mathbf{x} \in \Omega_h} \left| \mathcal{G}( \phi_t)(\mathbf{x}) - [g^{-1} \cdot \phi_{t+1} g ](\mathbf{x}) \right|^2,
\end{align}
where $\Omega_h \subset \mathbb{Z}^2$ denotes the discrete spatial grid. Here, $g \cdot \phi$ denotes the action of a group element $g \in D_4$ on the field $\phi$, which applies the corresponding spatial rotation or reflection. This action ensures consistent behavior under group operations. The dihedral group $D_4$ is defined as
\begin{equation}
D_4 = \{I, r, r^2, r^3, s, sr, sr^2, sr^3\},
\end{equation}
where $I$ is the identity transformation, $r$ represents a counterclockwise 90-degree rotation, and $s$ denotes a reflection operation on the 2D physical field. All other elements of the group are compositions of these basic symmetries. By minimizing $\mathcal{L}_{\text{eq}}$, the U-shaped Neural Operator is constructed to respect the equivariance property of the underlying physical process, leading to more robust and physically consistent predictions.

\begin{figure}[H]
\centering
\includegraphics[width=0.8\textwidth]{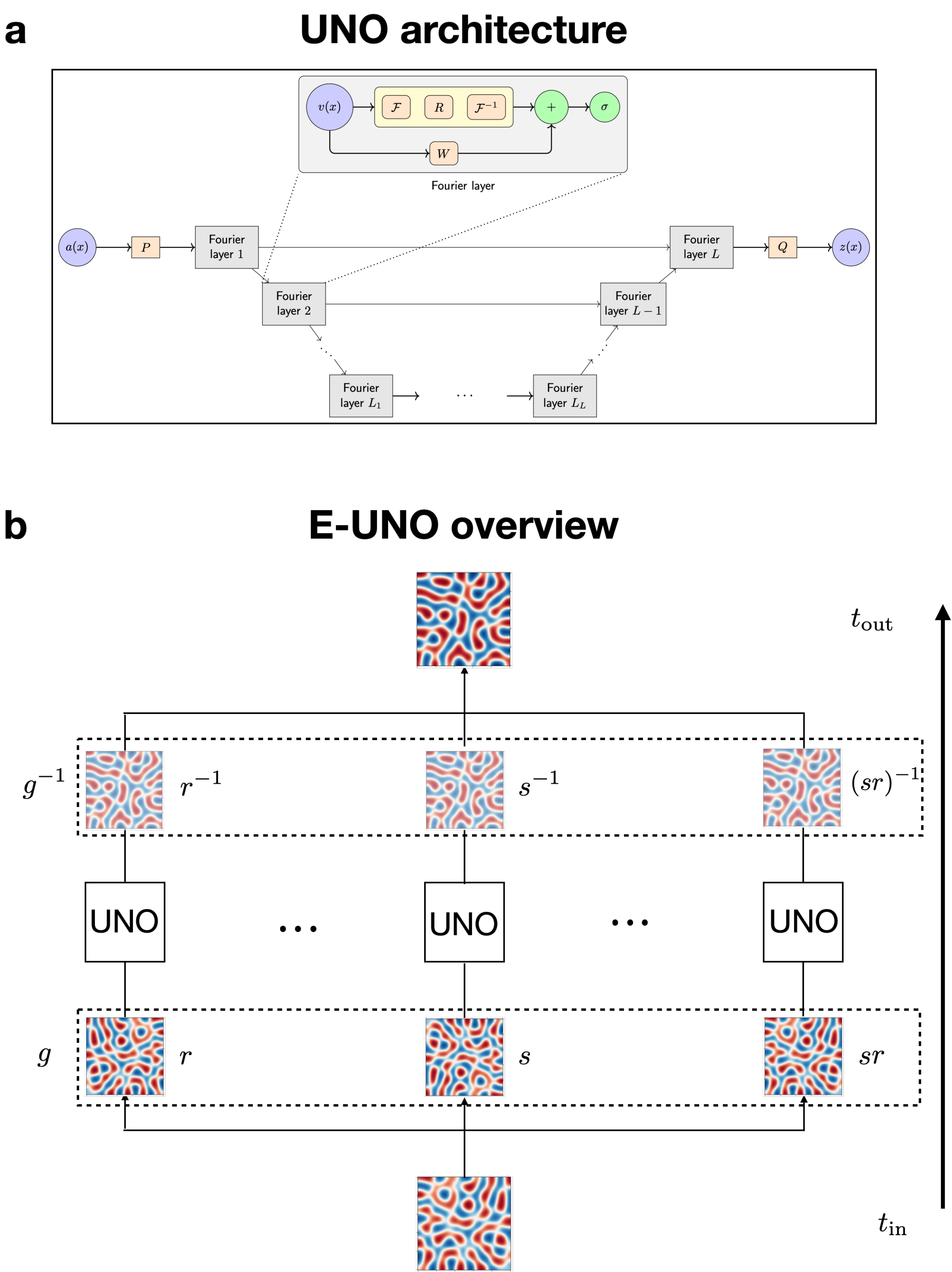}
\caption{Architectures of the U-shaped Neural Operator and its Equivariant Extension (E-UNO). Panel \textbf{a} Schematic of the UNO architecture, which combines Fourier layers for global spectral learning with a multi-resolution encoder–decoder structure. The input function $a(x)$ is lifted to a high-dimensional feature space through a projection $P$, processed through a sequence of Fourier layers, and finally projected back to the target output space by $Q$. Panel \textbf{b} Overview of the Equivariant UNO framework. The model enforces $D_4$-equivariance (rotations and reflections of a square domain) by applying group transformations $g \in D_4$ to the input, propagating through UNO, and then applying the corresponding inverse transformations to the outputs before loss evaluation. This ensures that predictions are consistent under all symmetry operations of the underlying physical system.}
\label{fig:E-UNO}
\end{figure}
Figure~\ref{fig:E-UNO} provides an overview of the baseline UNO and its equivariant extension (E-UNO). The UNO architecture (Fig.~\ref{fig:E-UNO}a) integrates global spectral convolution through Fourier layers with a multi-resolution encoder–decoder pathway, enabling the capture of both long-range dependencies and localized features in the phase-field evolution. The input field $a(x)$ is first lifted to a high-dimensional latent space by the projection operator $P$, processed through a sequence of Fourier layers that mix information across scales, and finally projected back to the physical output space $z(x)$ via $Q$.  
In the E-UNO framework (Fig.~\ref{fig:E-UNO}b), $D_4$-equivariance is enforced by applying each group transformation $g \in D_4$ (rotations and reflections of the square computational domain) to the input, propagating the transformed field through the UNO backbone, and applying the corresponding inverse transformation $g^{-1}$ to the output before loss computation. This procedure implements the equivariance constraint given in Eq.~(22), ensuring that the model’s predictions are consistent under all symmetry operations inherent to the governing equations. By embedding these physical symmetries into the architecture, E-UNO improves prediction stability, reduces data redundancy, and enhances generalization across diverse initial conditions and domain orientations.

\subsection{Loss function}

The loss function is composed of a data loss and the equivariance loss:
\begin{align}
    \mathcal{L} = \mathcal{L}_{\rm data} + \mathcal{L}_{\rm eq}.
\end{align}
For the data loss, two formulations are considered: $\mathcal{L}_{\rm data}=\mathcal{L}_{L_2}$ and $\mathcal{L}_{\rm data}=\mathcal{L}_{H_1}$ with
\begin{subequations}\label{eq:loss}
\begin{align}
\mathcal{L}_{L_2} = &~ \|\Phi - \hat{\Phi}\|^2_{L_2(\Omega_h)}, \\
\mathcal{L}_{H_1} =&~ \|\Phi - \hat{\Phi}\|_{H^1(\Omega_h)}= \|\Phi - \hat{\Phi}\|^2_{L_2(\Omega_h)} + \|\nabla (\Phi - \hat{\Phi})\|^2_{L_2(\Omega_h)},
\end{align}
\end{subequations}
where $\hat{\Phi}$ denotes the model prediction and $\Omega$ is the spatial domain. The $L_2$ loss measures only the difference between $\Phi$ and $\hat{\Phi}$, while the $H_1$ loss additionally incorporates differences between $\nabla \Phi$ and $\nabla \hat{\Phi}$.
\section{Results and discussions} \label{sec:results}
In this section, we systematically evaluate the performance of neural operator models for predicting microstructural evolution governed by the Cahn–Hilliard equation. We begin by describing the setup and data configuration. Subsequently, we validate the baseline UNO architecture, and afterwards demonstrate its ability to capture the spatiotemporal evolution of the phase-field and the corresponding behavior. We then examine how the number of input and output sequence steps affects prediction accuracy, highlighting the trade-off between temporal context and forecast horizon. To improve predictions in regions with sharp interfaces, we investigate the impact of incorporating spatial gradient information into the loss function, which enhances both accuracy and robustness across different initial conditions. 

After establishing this baseline, we evaluate the benefits of incorporating physical symmetries through our proposed Equivariant UNO (E-UNO) model. Comparative results show that enforcing D4-equivariance leads to systematic improvements over the standard UNO, and also outperforms the FNO. To further assess physical consistency, we compare the predicted free energy trajectories, confirming that E-UNO preserves thermodynamic trends and matches well with the ground truth results. Lastly, we test the model’s generalization capabilities by transferring it to super resolutions, demonstrating its potential for mesh-independent, super-resolved inference.

\subsection{Problem setup and data configuration}

The two-dimensional microstructural evolution of a binary material system, governed by~\cref{eq:Cahn}, serves as the testbed for model evaluation. The computational domain is a unit square discretized into a $100 \times 100$ uniform grid, giving a spatial resolution of $0.01\,\text{m}$ per cell. All simulations employ constant mobility $\gamma = 1$ in~\cref{eq:PDE} and interface thickness $\lambda = 0.01$ in~\cref{eq: free energy}.  

The dataset comprises 300 independent simulations performed with COMSOL Multiphysics. Each case starts from a uniform initial condition $\Phi = 0$, with variability introduced by a distinct random seed controlling the stochastic terms in the dynamics. This ensures that, despite identical initial states, each run evolves along a unique trajectory.  

From each simulation, we extract 30 temporal fragments, focusing on intervals of rapid interfacial evolution. To assess temporal generalization, models are trained with various configurations of input sequence length $n_{\text{in}}$ and output sequence length $n_{\text{out}}$, while maintaining a fixed sampling interval $\Delta t$. For the UNO parameter exploration, all models are trained for 50 epochs, balancing convergence with computational efficiency.

\begin{figure}[h!]
\centering
\includegraphics[width=0.6\textwidth]{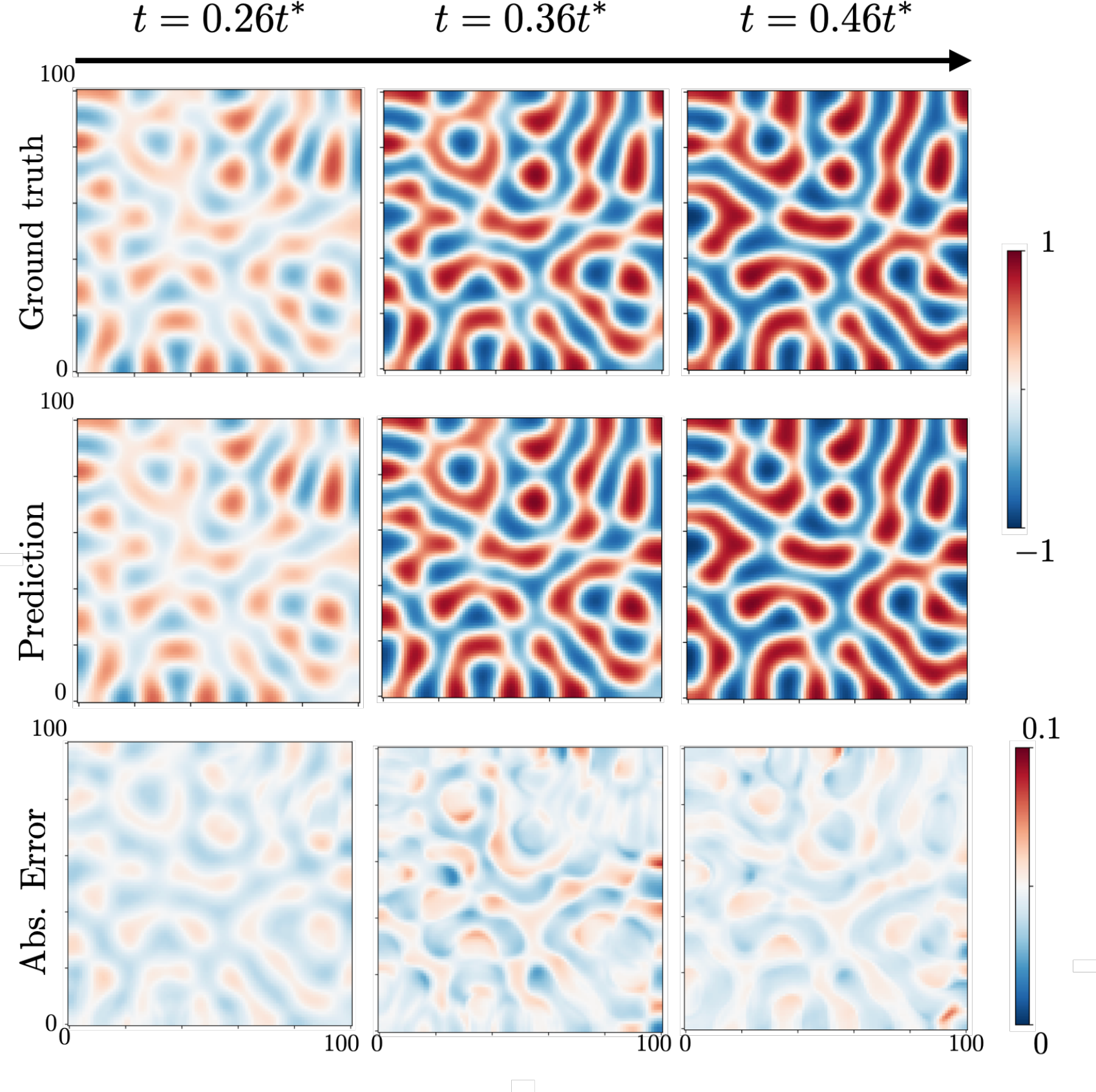}
\caption{Spatiotemporal evolution of the predicted field compared to the ground truth at three dimensionless time instances ($t = 0.26t^*$, $0.36t^*$, and $0.46t^*$). The top row shows the ground truth, the middle row displays the model predictions, and the bottom row presents the pointwise prediction error. The model accurately captures the complex spatial patterns with low prediction error across time. Colorbars indicate the field values (top two rows) and error magnitude (bottom row).} 
\label{fig:profile_pred}
\end{figure}
\subsection{Prediction of phase-field dynamics}
To establish a baseline for neural operator performance in modeling phase-field dynamics, we first employ the U-shaped Neural Operator model. This architecture captures the temporal evolution of the order parameter, $\hat{\Phi}$, and implicitly reflects the progression of free energy over time. We assess its performance against ground truth data generated from high-fidelity numerical simulations, which are performed over a total physical time of 0.3 second. For model training and prediction, we extract data at a sampling interval of $\Delta t = 0.01$ s, and use the first $n_{\text{in}} = 5$ frames as input. Each reference simulation requires approximately 240 seconds of CPU time. 

We divide the 300 simulation datasets into training, validation, and test sets using an 80/10/10 split to ensure reliable evaluation of model performance. The model’s predictive capability is illustrated at a sampling interval of $\Delta t$, where we present instantaneous snapshots from the test set across several dimensionless time instances, defined as $t^* = t / t_{\text{end}}$, with $t_{\text{end}}$ denoting the total simulation duration. The top row shows the ground truth evolution of the order parameter $\Phi$, while the second row displays the corresponding model predictions. The bottom row visualizes the absolute prediction error, at each time frame. The results show strong agreement between predictions and ground truth, with max absolute errors remaining within 5\%, indicating that the UNO model reliably captures the phase-field dynamics governed by the Cahn–Hilliard equation.

Overall, the UNO model achieves high accuracy in temporal prediction. For the entire process, inference requires only 0.03 seconds (See \cref{tab:inference_times}), whereas the baseline numerical simulation takes 240 seconds. This corresponds to a nominal speedup of approximately $8000$ times, underscoring the potential of neural operators to accelerate high-dimensional, time-dependent PDE simulations. We note, however, that this comparison is not entirely hardware-neutral: the ground truth reference was computed on a single CPU core, while inference was performed on an NVIDIA A100 GPU. Although this limits a strict one-to-one comparison, the results nonetheless illustrate the significant acceleration achievable when replacing conventional PDE solvers with data-driven operator surrogates.

\begin{figure}[h]
\centering
\includegraphics[width=1\textwidth]{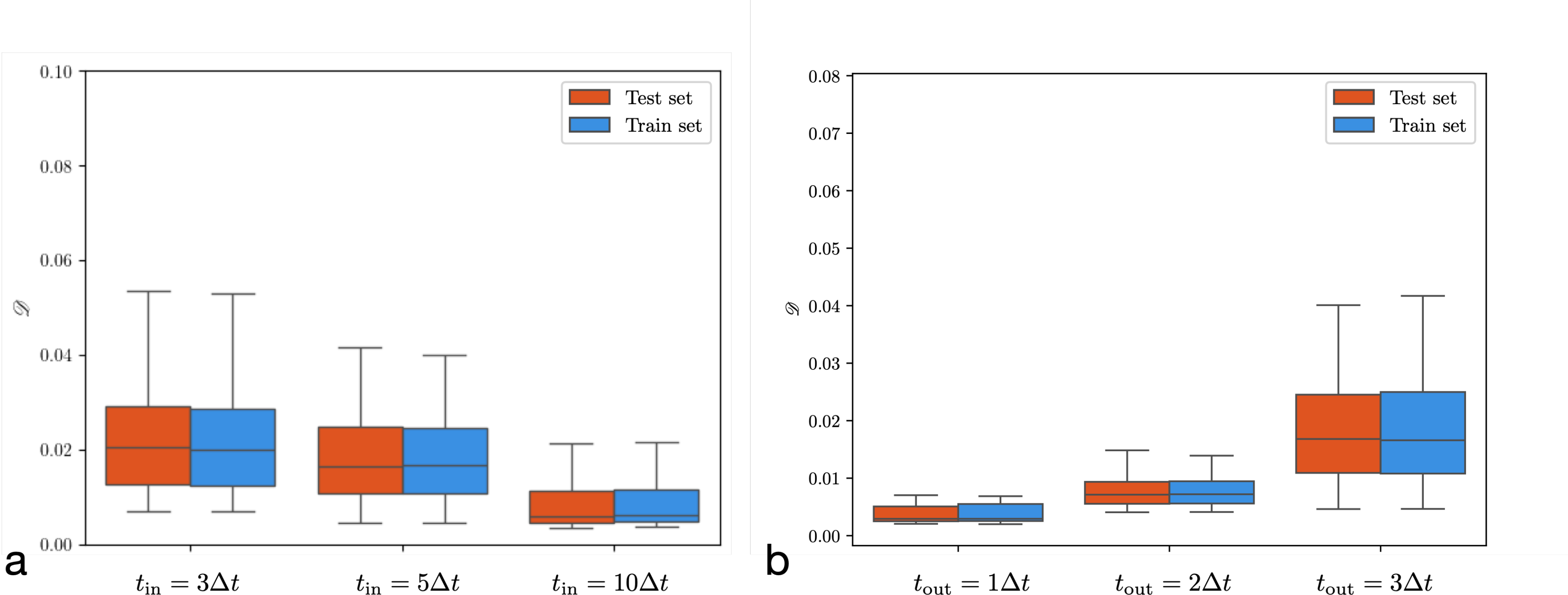}
\caption{Panel \textbf{a} Box plots of the error $\mathscr{D}$ for training and test sets at three time steps ($t_{out} = 1\Delta t$, $2\Delta t$, and $3\Delta t$). The boxes represent the interquartile range, the horizontal line inside each box shows the median, and whiskers indicate the range of the data excluding outliers. The results show increasing error variability with time, reflecting the growing prediction challenge as the temporal horizon extends. Panel \textbf{b} Box plots showing the distribution of the error $\mathscr{D}$ for training and test sets at three different input times ($t_{in} = 3\Delta t$, $5\Delta t$, and $10\Delta t$). The boxes represent the interquartile range, the central line indicates the median, and the whiskers extend to show the full spread of the data, excluding outliers. The comparison illustrates how the error decreases with increasing input time.}
\label{fig:boxplot_output}
\end{figure}

\subsection{Influence of temporal sequence length on model accuracy}
\label{sec:input_output_analysis}

We investigate two key hyperparameters in the prediction strategy of the neural operator: the number of input frames, $t_{\text{in}} = n_{\text{in}}\Delta t$, and the number of output frames, $t_{\text{out}} = n_{\text{out}}\Delta t$. These parameters play distinct roles in controlling model behavior. A larger $n_{\text{in}}$ provides the model with more temporal context from the past, while increasing $n_{\text{out}}$ extends the prediction horizon, inherently making the forecasting task more challenging due to error accumulation.

To quantify predictive performance, we compute the spatially integrated relative $L_2$ error at each time step, defined as:
\begin{equation}
    \mathscr{D}(t) = \frac{\|\Phi(t) - \hat{\Phi}(t)\|_2^2}{\|\Phi(t)\|_2^2}, 
\end{equation}
where we suppress dependency on the spatial coordinates. This metric is evaluated across the entire evolution process for both the training and test datasets.

\paragraph{Effect of input time sequence length}
We first assess the influence of the input sequence length by fixing the prediction horizon at $t_{\text{out}} = 3\Delta t$, which represents the most challenging case due to its longer extrapolation window. The results, presented in Fig.~\ref{fig:boxplot_output}a, show that increasing the number of input frames leads to improved prediction accuracy. In particular, increasing $t_{\text{in}}$ from $5\Delta t$ to $10\Delta t$ results in a notable reduction in error, indicating that access to a longer temporal history helps the model better capture the underlying dynamics. Despite this, all configurations maintain mean errors below 0.02, underscoring the overall robustness of the UNO model.

\paragraph{Effect of output time sequence length}
Next, we fix the number of input frames at $t_{\text{in}} = 5\Delta t$ and evaluate how extending the prediction horizon affects model performance. As shown in Fig.~\ref{fig:boxplot_output}b, prediction error increases with $t_{\text{out}}$. This is especially pronounced at $t_{\text{out}} = 3\Delta t$, where accumulated uncertainties make accurate prediction more difficult. This observation is consistent with prior findings~\cite{Li2020FourierNO} that highlight the compounding nature of forecast error in long-horizon autoregressive prediction. While UNO performs well for short-term forecasting, its predictive accuracy degrades as the output sequence length increases.

\paragraph{$H^1$ loss enhancement} We also explored the use of a gradient-enhanced loss function to improve spatial accuracy near interfacial regions. Compared to the standard training objective, incorporating the gradient of the order parameter led to a ~37\% reduction in early-stage prediction error (peak error reduced from 6\% to 3.8\%), and reduced overall error variance across different initial conditions. These improvements are most evident during the rapid phase separation stage. Detailed comparisons are provided in~\ref{app:grad}.

\begin{figure}[H]
    \centering
    \includegraphics[width=1\linewidth]{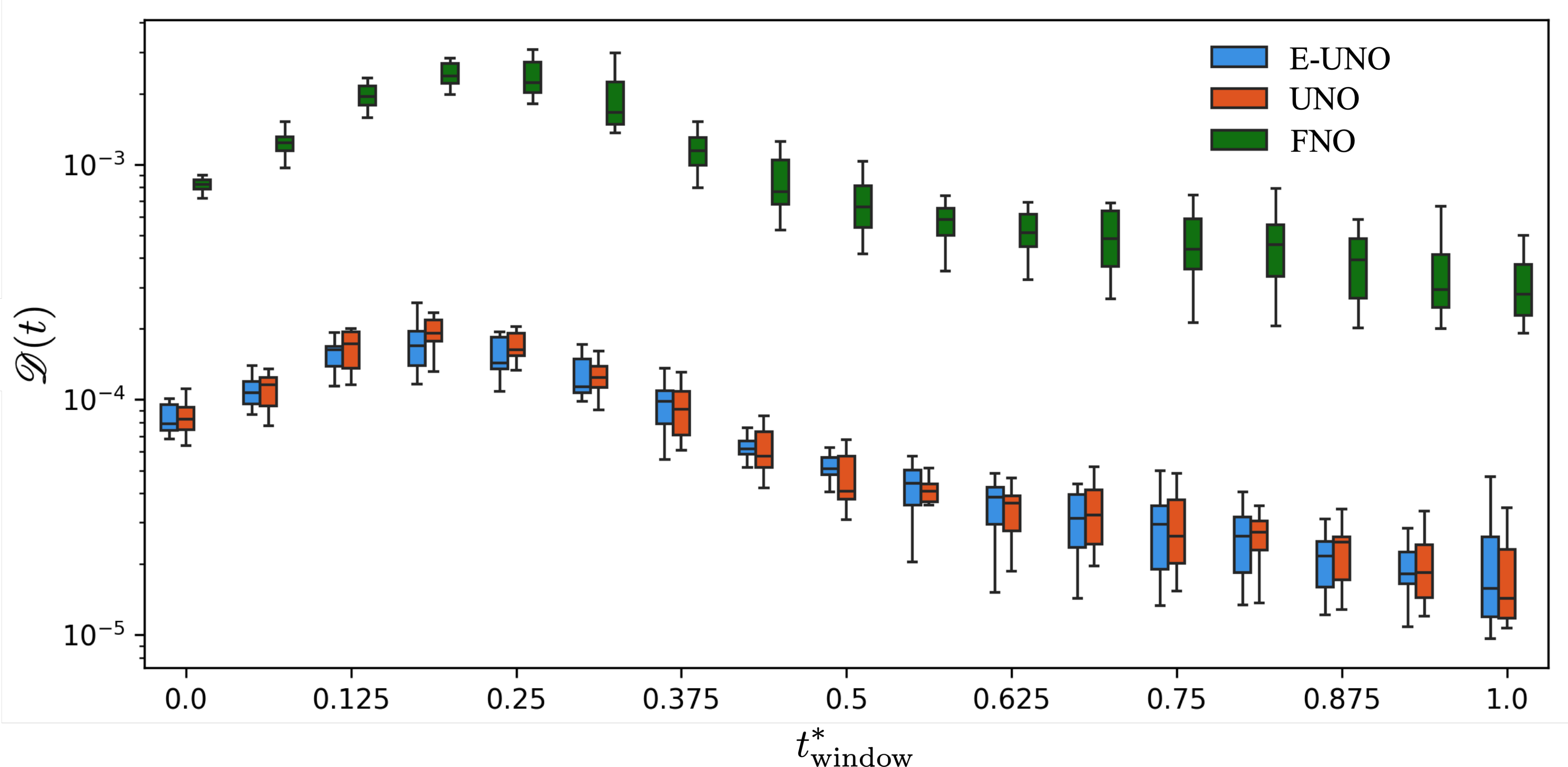}
    \caption{
    Temporal evolution of the spatially integrated relative $L_2$ error $\mathcal{D}(t)$ for E-UNO (blue), UNO (orange), and FNO (green) across normalized time windows $t_{\text{window}}^*$. Each $t_{\text{window}}^*$ value represents the error averaged over a time window across three-step prediction sequences for all validation cases. The boxes indicate the interquartile range (25th–75th percentile) of the error distribution, while the horizontal line inside each box marks the median. Whiskers denote the full range of observed errors, capturing the variability across predictions.
    }
    \label{fig:eq_loss_comparison}
\end{figure}
\subsection{Equivariant U-shaped neural operators comparison with other neural operators}
Building upon the findings in Section~\ref{sec:input_output_analysis}, where longer input sequences and gradient-enhanced losses improved the baseline UNO’s accuracy but could not address its lack of symmetry awareness, we now incorporate $D_4$-equivariance directly into the architecture. The resulting Equivariant U-shaped Neural Operator is designed to respect the rotational and reflectional symmetries of the square computational domain, thereby enhancing robustness and reducing redundant learning. We first compare E-UNO with the baseline UNO to quantify the benefits of symmetry enforcement. Based on previous study, we used 5 input timeframes so that it won't ignore the early stage evolution of the model, and predict 3 $t_{\text{out}}$ iteratively. To make a fair comparison, all models for E-UNO, UNO, FNO and equivariance informed FNO are performed over 200 epochs of training and select the best of the validation loss. 

Figure~\ref{fig:eq_loss_comparison} compares the temporal evolution of the spatially integrated relative $L_2$ error, $\mathcal{D}(t)$, for E-UNO, UNO, and FNO over normalized time windows $t_{\text{window}}^*$. Here, $t_{\text{window}}^*$ denotes the time-averaged error within each window, computed over three-step prediction sequences for all validation cases. Across all stages of the evolution, both E-UNO and UNO consistently achieve errors an order of magnitude lower than FNO. The error distributions for E-UNO and UNO remain tightly clustered, as reflected by the narrow interquartile ranges, while FNO exhibits substantially higher median errors and wider spreads. The benefits of symmetry enforcement are evident in the lower median and reduced variability of E-UNO compared to UNO, particularly in the early- to mid-stage evolution where morphological changes are most pronounced. These results confirm that enforcing $D_4$-equivariance yields the greatest benefits in dynamic regimes with high morphological activity, where symmetry constraints guide the network toward physically consistent predictions. Quantitatively, E-UNO achieves a consistent reduction in prediction error across the evolution process. Averaged over all time windows, the max relative $L_2$ error $\mathscr{D}(t)$ is reduced by 34.32\% compared to UNO at early stages where the evolution is dramatic. The improvement is most pronounced during the early and intermediate stages of the simulation ($t_{\text{window}} \leq 0.4t^*$). The median error improvement is up to 11\%. These gains indicate not only a reduction in bias but also improved stability across spatial locations and test samples. In the late stages of evolution, where microstructural coarsening slows and the system approaches equilibrium, the performance of the two models converges, suggesting that the primary advantage of equivariance lies in accurately capturing dynamic, symmetry-rich regimes. 

\begin{figure}[h!]
    \centering
    \includegraphics[width=1\linewidth]{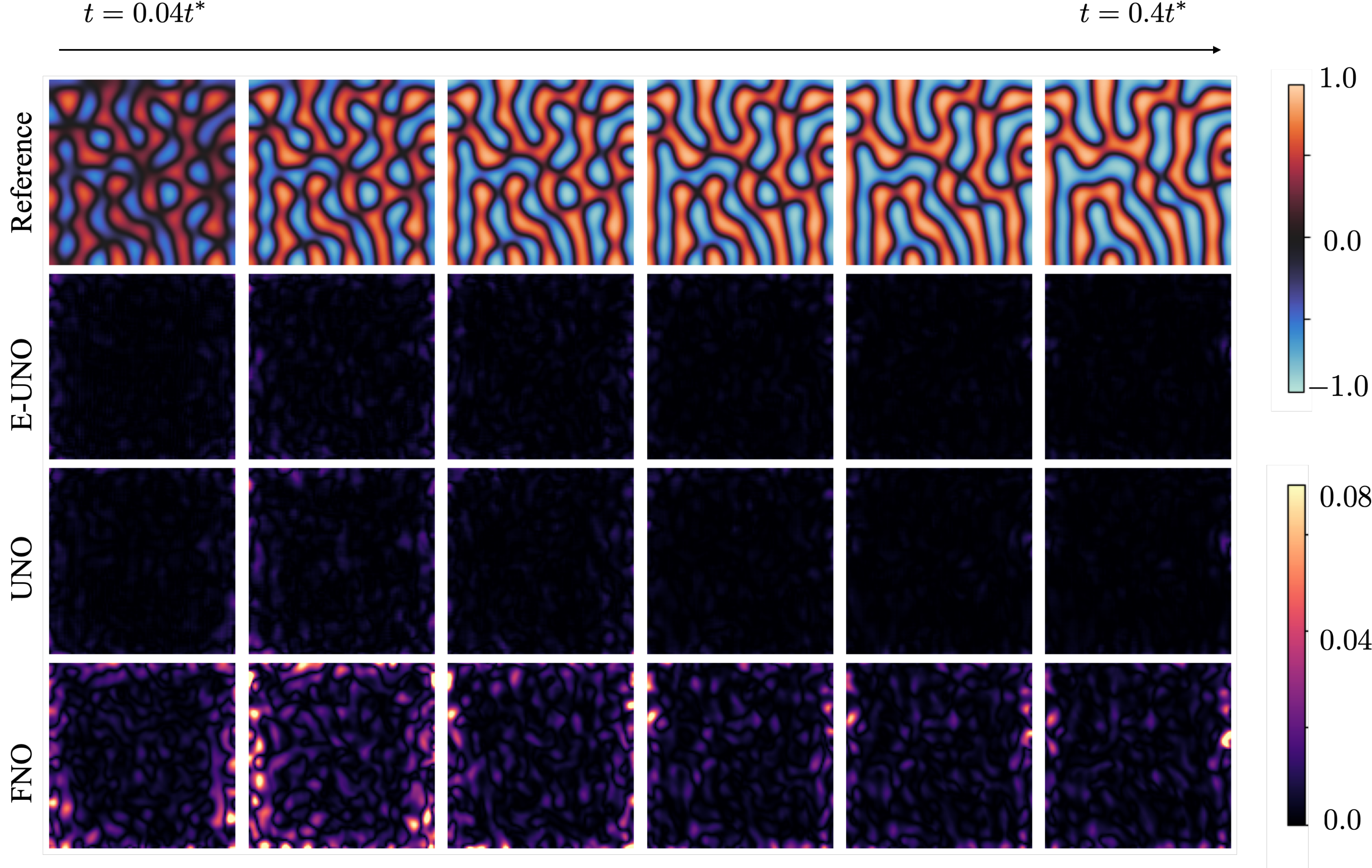}
    \caption{
        Comparison of predicted phase-field evolution errors for E-UNO, UNO, and FNO from $t = 0.04t^*$ to $t = 0.4t^*$. The top row shows reference ground-truth fields, while the lower rows present the absolute prediction error for each model, with separate color scales for the phase field (top) and error maps (bottom three rows). E-UNO exhibits the smallest and most spatially uniform error across all time frames. Both E-UNO and UNO unrecognizable error, and FNO displays substantially larger and more structured error patterns, particularly near phase interfaces.
    }
    \label{fig:E-UNO_evolution}
\end{figure}

We exam the evolution of the dynamics by comparison the spatio error at each snapshot for each neural operator. Figure~\ref{fig:E-UNO_evolution} illustrates the spatial distribution of absolute prediction errors for E-UNO, UNO, and FNO over an early to mid-stage evolution window ($t = 0.04t^*$ to $t = 0.40t^*$). The ground-truth reference fields (top row) depict the rapid morphological changes characteristic of this phase separation stage. E-UNO achieves near-uniformly low errors across the domain, with minimal localized deviations at interfaces. In contrast, UNO produces slightly higher interface-associated errors, while FNO exhibits significantly larger and more spatially correlated error structures, particularly along evolving phase boundaries. 

We also evaluate the impact of equivariance on the FNO architecture, and the results are consistent with our observations for the UNO architecture, confirming that the benefits of symmetry enforcement are not architecture-specific (see~\ref{app:fno_equivariance}). For the same parameter settings, the FNO baseline exhibits errors roughly an order of magnitude larger than those of UNO, reflecting the advantage of the U-shaped encoder–decoder design in capturing multi-scale dynamics. For completeness, we provide a consolidated comparison of E-UNO, UNO, E-FNO, and FNO in terms of both median prediction errors and inference speeds, showing that while E-UNO achieves the lowest errors overall, it retains the same inference times as its non-equivariant counterpart (see~\ref{sec:system}).

\subsection{Free energy evolution with E-UNO}
\label{sec:E-UNO_free_energy}

We now evaluate the physical fidelity of the proposed E-UNO model by examining its ability to capture the temporal evolution of the normalized free energy $F/F_0$ in the Cahn–Hilliard system. Here the focus is on assessing whether the explicit enforcement of $D_4$-equivariance can learn the ground truth thermodynamic trends.
\begin{figure}[h!]
\centering
\includegraphics[width=0.8\textwidth]{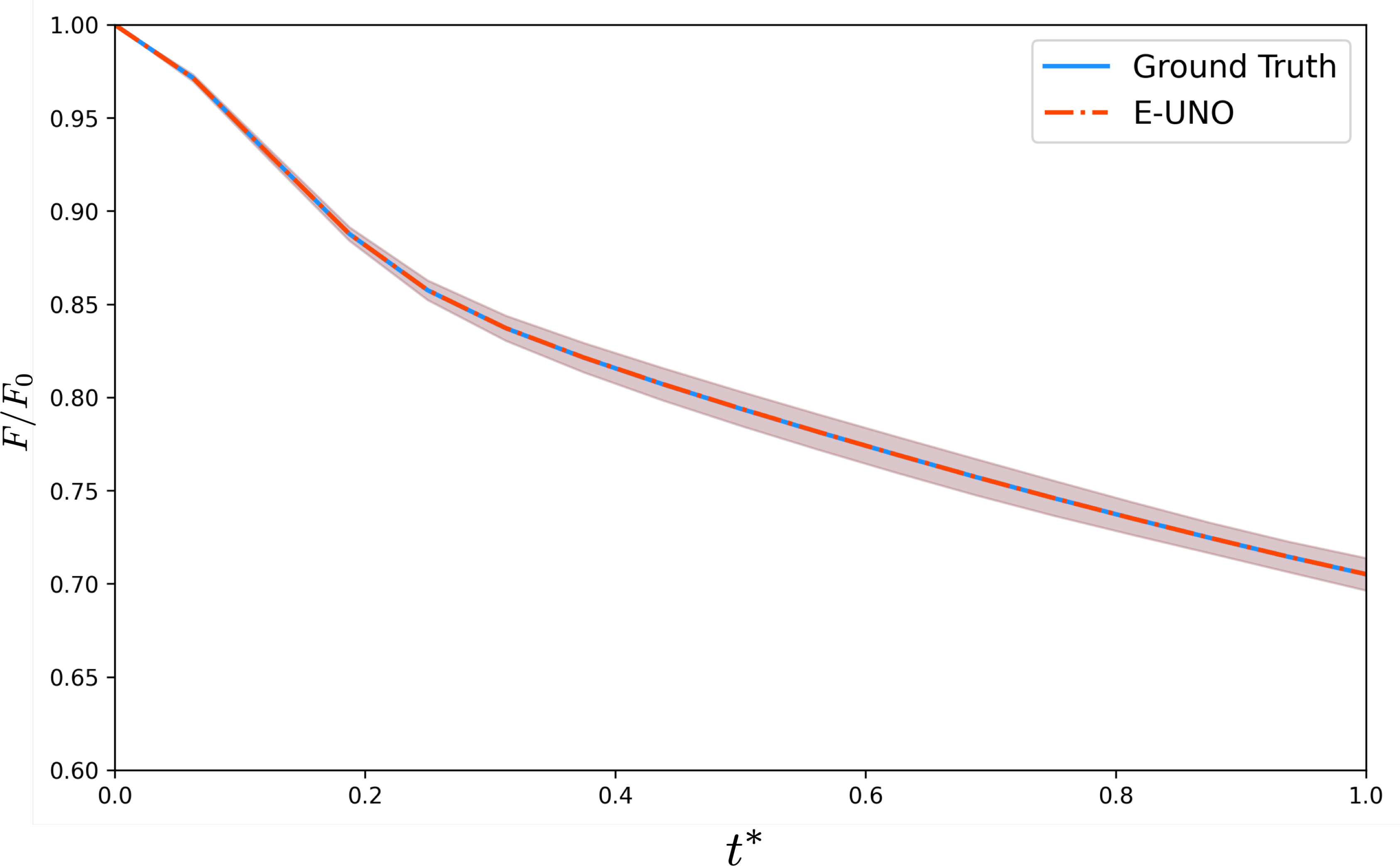}
\caption{Temporal evolution of the normalized free energy, $F/F_0$, predicted by E-UNO (red dash–dot) compared with ground truth numerical simulation (blue solid). The shaded region denotes the range of predicted values across all validation cases. E-UNO closely follows the reference curve throughout the entire evolution, with differences barely distinguishable at the plotting scale.} 
\label{fig:free_energy_E-UNO}
\end{figure}

Figure~\ref{fig:free_energy_E-UNO} presents a systematic comparison of the normalized free energy $F/F_0$ obtained from E-UNO predictions against the ground truth simulation data. Across the entire evolution, the E-UNO curve is almost perfectly superimposed on the reference, indicating excellent agreement in capturing the thermodynamic decay trend. The shaded band, representing the spread of predictions over all validation cases, remains narrow throughout, demonstrating both accuracy and consistency. The close match suggests that E-UNO not only reproduces the spatial morphology of the phase field but also preserves the underlying energy dissipation dynamics dictated by the Cahn–Hilliard model. The agreement of E-UNO with the ground truth decay curve therefore confirms that embedding $D_4$ symmetry enhances the model’s ability to learn physically meaningful representations.

\subsection{Super-resolution generalization with E-UNO}
\begin{figure}
\centering
\includegraphics[width=1\textwidth]{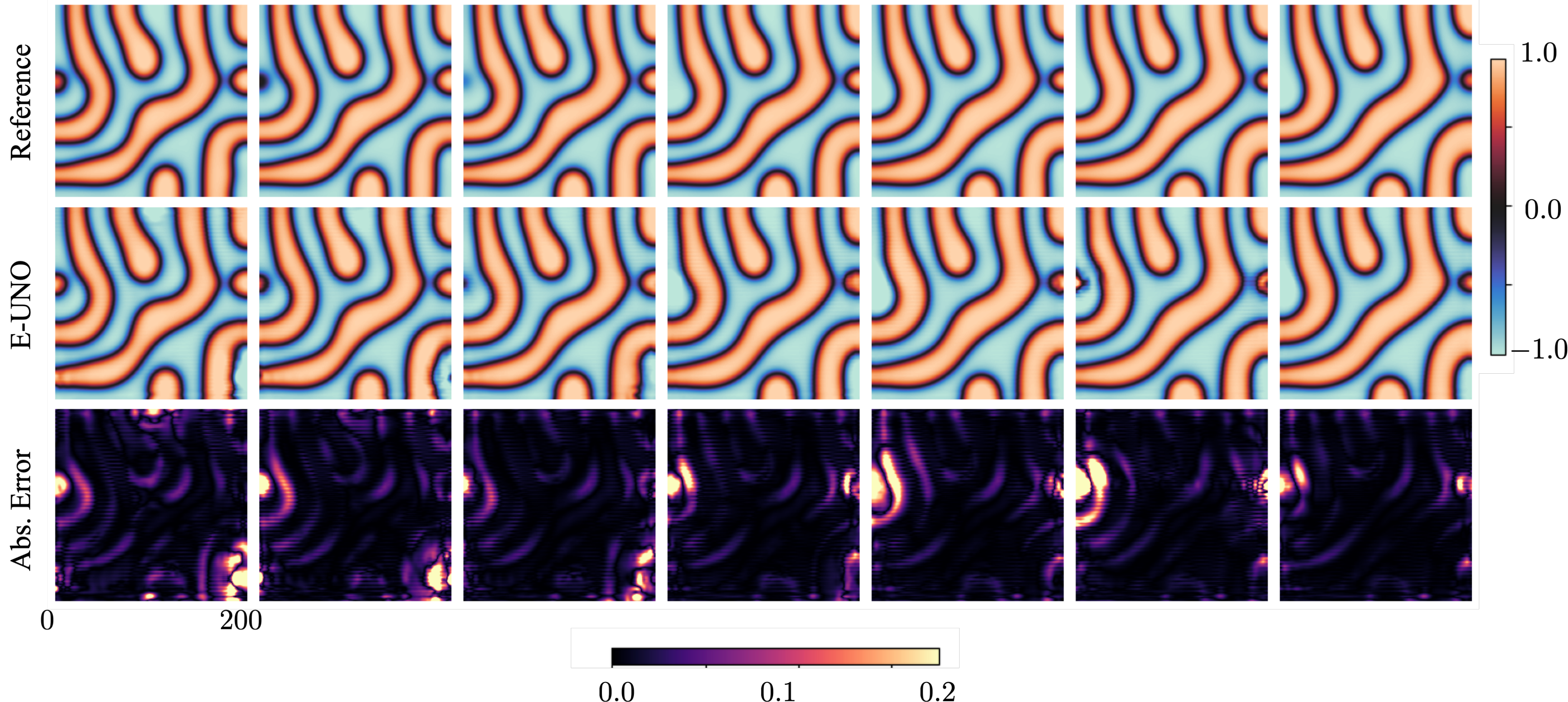}
\caption{Super-resolution prediction of phase-field evolution using E-UNO, compared to the ground truth reference, for $t=0.5t^*$ to $t=1.0t^*$. The E-UNO is trained on coarse $100\times 100$ data and evaluated directly on a refined $200\times 200$ grid without retraining. The top row shows the ground truth reference fields, the middle row shows E-UNO predictions, and the bottom row displays the absolute error. 
E-UNO accurately resolves fine-scale interfacial structures and preserves morphological patterns across the domain, with most errors remaining below 0.1.}
\label{fig:super_resolution}
\end{figure}
A desirable property of E-UNO is the ability to generalize across spatial resolutions without retraining. To evaluate the model’s super-resolution capability, we trained it on a coarse $100 \times 100$ grid using 5 input timestep of $t_{in}=5\Delta t$ and 3 predictive output steps, and then directly applied it to a finer $200 \times 200$ grid without any additional fine-tuning. It is important to note that, due to the randomness in the initialization process, the initial conditions at low resolution cannot be exactly reproduced at higher resolutions. Despite this, the model demonstrates strong generalization. 
Figure~\ref{fig:super_resolution} demonstrates the super-resolution capability of E-UNO, where the model predicts times ranging from $t=0.5t^*$ to $t=1.0t^*$. The predictions closely match the high-resolution reference, accurately capturing both large-scale morphological features and fine-scale interfacial details. The absolute error maps confirm that max discrepancies remain localized and below 10\% in magnitude, indicating that E-UNO retains strong mesh-independent generalization. This confirms that the incorporation of $D_4$-equivariance does not compromise the super-resolution performance.

\section{Conclusion}\label{sec:conclusion}
In this work, we developed the Equivariant U-shaped Neural Operator, a symmetry-aware operator-learning framework for predicting microstructural evolution governed by the Cahn–Hilliard equation. The method enhances $D_4$-equivariance by applying group transformations to the input fields before inference and the corresponding inverse transformations to the outputs before loss evaluation. This approach ensures that the model’s predictions remain consistent under all rotations and reflections of the square computational domain, without modifying the internal architecture of the baseline U-shaped Neural Operator. Systematic evaluations demonstrated several key advantages over both the non-equivariant U-shaped Neural Operator and the Fourier Neural Operator. First, the proposed model consistently reduced pointwise prediction errors by up to 34\% in early and intermediate evolution stages, where interface dynamics are most active. Second, it preserved global thermodynamic consistency, achieving an approximately 11\% reduction in free energy deviation relative to the U-shaped Neural Operator. Third, it maintained strong mesh-independent generalization capability, accurately performing super-resolution predictions without retraining. These benefits were achieved without compromising inference speed, confirming that symmetry enforcement through external group operations offers an effective and lightweight enhancement to existing operator-learning models. Overall, this study underscores the synergy between the phase-field method and symmetry-aware operator learning. The proposed framework, built upon the efficiency and scalability of the U-shaped Neural Operator, demonstrates that enforcing $D_4$-equivariance through pre- and post-processing operations can significantly enhance accuracy and physical fidelity without increasing computational cost. Its ability to generalize across resolutions, preserve thermodynamic trends, and respect underlying physical symmetries points toward integrating symmetry-enforced, data-driven operators into traditional computational frameworks to enable real-time or large-scale phase-field predictions. This work contributes to bridging the gap between physics-based modeling and machine learning, paving the way for more robust, efficient, and physically consistent machine-learning–accelerated simulation tools. Future research directions include the extension to different phase-field models, e.g. $N$-phase Cahn-Hilliard (mixture) models \cite{eyre1993systems,li2016multi,ten2024thermodynamically,ten2025unified,ten2025compressible}, Cahn-Hilliard with elasticity \cite{garcke2003cahn,oudich2025phase}, and Cahn-Hilliard with crystal growth \cite{golovin1998convective,balakrishna2018combining}, as well as improving the neural operator architecture, e.g. through more efficiently incorporating the $D_4$-equivariance component.

\section{Acknowledgements}
P.V.C. and X.X. acknowledge funding support from European Commission CompBioMed Centre of Excellence (Grant No. 675451 and 823712). Support from the UK Engineering and Physical Sciences Research Council under the projects ``UK Consortium on Mesoscale Engineering Sciences (UKCOMES)" (Grant No.
EP/R029598/1) and ``Software Environment for Actionable and VVUQ-evaluated Exascale Applications (SEAVEA)" (Grant No. EP/W007711/1) is gratefully acknowledged. P.V.C. and X.X. acknowledge 2024-2025 DOE INCITE award for computational resources on supercomputers at the Oak Ridge Leadership Computing Facility under the ``COMPBIO3" project. P.V.C. and X.X. acknowledge the use of resources provided by the Isambard-AI National AI Research Resource (AIRR). Isambard-AI is operated by the University of Bristol and is funded by the UK Government’s Department for Science, Innovation and Technology (DSIT) via UK Research and Innovation; and the Science and Technology Facilities Council [ST/AIRR/I-A-I/1023].

\bibliographystyle{unsrt}
\bibliography{reference}

\begin{thebibliography}{10}

\bibitem{anderson1998diffuse}
D.M. Anderson, G.B. McFadden, and A.A. Wheeler.
\newblock Diffuse-interface methods in fluid mechanics.
\newblock {\em Annual review of fluid mechanics}, 30:139--165, 1998.

\bibitem{kim2012phase}
J.~Kim.
\newblock Phase-field models for multi-component fluid flows.
\newblock {\em Communications in Computational Physics}, 12:613--661, 2012.

\bibitem{wight2020solving}
C.L. Wight and J.~Zhao.
\newblock {Solving Allen-Cahn and Cahn-Hilliard equations using the adaptive
  physics informed neural networks}.
\newblock {\em arXiv preprint arXiv:2007.04542}, 2020.

\bibitem{vanderWaals1979}
J.~D. van~der Waals.
\newblock The thermodynamic theory of capillarity under the hypothesis of a
  continuous variation of density.
\newblock {\em Journal of Statistical Physics}, 20:200--244, 1979.

\bibitem{cahn1958free}
J.W. Cahn and J.E. Hilliard.
\newblock Free energy of a nonuniform system. {I}. {I}nterfacial free energy.
\newblock {\em The Journal of chemical physics}, 28:258--267, 1958.

\bibitem{gurtinmodel}
M.E. Gurtin, D.~Polignone, and J.~Vinals.
\newblock Two-phase binary fluids and immiscible fluids described by an order
  parameter.
\newblock {\em Math. Models Methods Appl. Sci.}, 6:815--831, 1996.

\bibitem{lowengrub1998quasi}
J.~Lowengrub and L.~Truskinovsky.
\newblock Quasi--incompressible cahn--hilliard fluids and topological
  transitions.
\newblock {\em Proceedings of the Royal Society of London. Series A:
  Mathematical, Physical and Engineering Sciences}, 454(1978):2617--2654, 1998.

\bibitem{ten2023unified}
M.F.P. ten Eikelder, K.G. Van Der~Zee, I.~Akkerman, and D.~Schillinger.
\newblock {A unified framework for Navier--Stokes Cahn--Hilliard models with
  non-matching densities}.
\newblock {\em Mathematical Models and Methods in Applied Sciences},
  33:175--221, 2023.

\bibitem{abels2024mathematical}
H.~Abels, H.~Garcke, and A.~Poiatti.
\newblock Mathematical analysis of a diffuse interface model for multi-phase
  flows of incompressible viscous fluids with different densities.
\newblock {\em Journal of Mathematical Fluid Mechanics}, 26:29, 2024.

\bibitem{ten2025unified}
M.F.P. ten Eikelder.
\newblock {A unified framework for N-phase Navier--Stokes Cahn--Hilliard
  Allen--Cahn mixture models with non-matching densities}.
\newblock {\em Journal of Fluid Mechanics}, 1013:A26, 2025.

\bibitem{knopf2025thermodynamically}
P.~Knopf and Y.~Liu.
\newblock A thermodynamically consistent free boundary model for two-phase
  flows in an evolving domain with moving contact lines, variable contact
  angles and bulk-surface interaction.
\newblock {\em arXiv preprint arXiv:2507.01618}, 2025.

\bibitem{ten2025compressible}
M.F.P. ten Eikelder, E.H. van Brummelen, and D.~Schillinger.
\newblock Compressible n-phase fluid mixture models.
\newblock {\em arXiv preprint arXiv:2503.24225}, 2025.

\bibitem{kay2007efficient}
D.~Kay and R.~Welford.
\newblock Efficient numerical solution of cahn--hilliard--navier--stokes fluids
  in 2d.
\newblock {\em SIAM Journal on Scientific Computing}, 29:2241--2257, 2007.

\bibitem{minjeaud2013unconditionally}
S.~Minjeaud.
\newblock {An unconditionally stable uncoupled scheme for a triphasic
  Cahn--Hilliard/Navier--Stokes model}.
\newblock {\em Numerical Methods for Partial Differential Equations},
  29:584--618, 2013.

\bibitem{ten2024divergence}
M.F.P. ten Eikelder and D.~Schillinger.
\newblock {The divergence-free velocity formulation of the consistent
  Navier-Stokes Cahn-Hilliard model with non-matching densities,
  divergence-conforming discretization, and benchmarks}.
\newblock {\em J. Comput. Phys.}, 513:113148, 2024.

\bibitem{garcke2020long}
H.~Garcke and S.~Yayla.
\newblock Long-time dynamics for a cahn--hilliard tumor growth model with
  chemotaxis.
\newblock {\em Zeitschrift f{\"u}r angewandte Mathematik und Physik}, 71:123,
  2020.

\bibitem{ebenbeck2019analysis}
M.~Ebenbeck and H.~Garcke.
\newblock {Analysis of a Cahn--Hilliard--Brinkman model for tumour growth with
  chemotaxis}.
\newblock {\em Journal of Differential Equations}, 266:5998--6036, 2019.

\bibitem{bertozzi2007analysis}
A.~Bertozzi, S.~Esedoglu, and A.~Gillette.
\newblock Analysis of a two-scale {C}ahn--{H}illiard model for binary image
  inpainting.
\newblock {\em Multiscale Modeling \& Simulation}, 6:913--936, 2007.

\bibitem{khain2008generalized}
E.~Khain and L.M. Sander.
\newblock Generalized {C}ahn-{H}illiard equation for biological applications.
\newblock {\em Physical Review E—Statistical, Nonlinear, and Soft Matter
  Physics}, 77:051129, 2008.

\bibitem{elson2010phase}
E.L. Elson, E.~Fried, J.E. Dolbow, and G.M. Genin.
\newblock Phase separation in biological membranes: integration of theory and
  experiment.
\newblock {\em Annual review of biophysics}, 39:207--226, 2010.

\bibitem{berry2018physical}
J.~Berry, C.P. Brangwynne, and M.~Haataja.
\newblock Physical principles of intracellular organization via active and
  passive phase transitions.
\newblock {\em Reports on Progress in Physics}, 81:046601, 2018.

\bibitem{brown2020language}
T.~Brown, B.~Mann, N.~Ryder, M.~Subbiah, J.~D. Kaplan, P.~Dhariwal,
  A.~Neelakantan, P.~Shyam, G.~Sastry, A.~Askell, et~al.
\newblock Language models are few-shot learners.
\newblock {\em Advances in neural information processing systems},
  33:1877--1901, 2020.

\bibitem{voulodimos2018deep}
A.~Voulodimos, N.~Doulamis, A.~Doulamis, and E.~Protopapadakis.
\newblock Deep learning for computer vision: A brief review.
\newblock {\em Computational intelligence and neuroscience}, 2018:7068349,
  2018.

\bibitem{ding2014data}
S.X. Ding.
\newblock {\em Data-driven design of fault diagnosis and fault-tolerant control
  systems}.
\newblock Springer, 2014.

\bibitem{brunton2020machine}
S.L. Brunton, B.R. Noack, and P.~Koumoutsakos.
\newblock Machine learning for fluid mechanics.
\newblock {\em Annual Review of Fluid Mechanics}, 52:477--508, 2020.

\bibitem{karniadakis2021physics}
G.E. Karniadakis, I.G. Kevrekidis, L.~Lu, P.~Perdikaris, S.~Wang, and L.~Yang.
\newblock Physics-informed machine learning.
\newblock {\em Nature Reviews Physics}, 3:422--440, 2021.

\bibitem{cheng2025machine}
S.~Cheng, M.~Bocquet, W.~Ding, T.S. Finn, R.~Fu, J.~Fu, Y.~Guo, E.~Johnson,
  S.~Li, C.~Liu, et~al.
\newblock Machine learning for modelling unstructured grid data in
  computational physics: a review.
\newblock {\em Information Fusion}, page 103255, 2025.

\bibitem{wang2025quantum}
M.~Wang, X.~Xue, M.~Gao, and P.V. Coveney.
\newblock {Quantum-informed machine learning for the prediction of chaotic
  dynamical systems}.
\newblock {\em arXiv preprint arXiv:2507.19861}, 2025.

\bibitem{fukami2019synthetic}
K.~Fukami, Y.~Nabae, K.~Kawai, and K.~Fukagata.
\newblock Synthetic turbulent inflow generator using machine learning.
\newblock {\em Physical Review Fluids}, 4:064603, 2019.

\bibitem{xue2022synthetic}
X.~Xue, H.-D. Yao, and L.~Davidson.
\newblock Synthetic turbulence generator for lattice boltzmann method at the
  interface between rans and les.
\newblock {\em Physics of Fluids}, 34:055118, 2022.

\bibitem{yang2019predictive}
X.~Yang, S.~Zafar, J.-X. Wang, and H.~Xiao.
\newblock Predictive large-eddy-simulation wall modeling via physics-informed
  neural networks.
\newblock {\em Physical Review Fluids}, 4:034602, 2019.

\bibitem{xue2024physics}
X.~Xue, S.~Wang, H.-D. Yao, L.~Davidson, and P.V. Coveney.
\newblock Physics informed data-driven near-wall modelling for lattice
  boltzmann simulation of high reynolds number turbulent flows.
\newblock {\em Communications Physics}, 7:338, 2024.

\bibitem{cheng2020data}
M.~Cheng, F.~Fang, C.~C. Pain, and I.M. Navon.
\newblock Data-driven modelling of nonlinear spatio-temporal fluid flows using
  a deep convolutional generative adversarial network.
\newblock {\em Computer Methods in Applied Mechanics and Engineering},
  365:113000, 2020.

\bibitem{Li2020FourierNO}
Z.-Y. Li, N.B. Kovachki, K.~Azizzadenesheli, B.~Liu, K.~Bhattacharya, A.M.
  Stuart, and A.~Anandkumar.
\newblock Fourier neural operator for parametric partial differential
  equations.
\newblock {\em ArXiv}, abs/2010.08895, 2020.

\bibitem{lu2021learning}
L.~Lu, P.~Jin, G.~Pang, Z.~Zhang, and G.E. Karniadakis.
\newblock Learning nonlinear operators via deeponet based on the universal
  approximation theorem of operators.
\newblock {\em Nature Machine Intelligence}, 3:218--229, 2021.

\bibitem{cao2024laplace}
Q.~Cao, S.~Goswami, and G.E. Karniadakis.
\newblock Laplace neural operator for solving differential equations.
\newblock {\em Nature Machine Intelligence}, 6:631--640, 2024.

\bibitem{shen2019phase}
Z.-H. Shen, J.-J. Wang, J.-Y. Jiang, S.X. Huang, Y.-H. Lin, C.-W. Nan, L.-Q.
  Chen, and Y.~Shen.
\newblock Phase-field modeling and machine learning of
  electric-thermal-mechanical breakdown of polymer-based dielectrics.
\newblock {\em Nature Communications}, 10:1843, 2019.

\bibitem{feng2021machine}
Y.~Feng, Q.~Wang, D.~Wu, Z.~Luo, X.~Chen, T.~Zhang, and W.~Gao.
\newblock Machine learning aided phase field method for fracture mechanics.
\newblock {\em International Journal of Engineering Science}, 169:103587, 2021.

\bibitem{TEICHERT2019201}
G.H. Teichert, A.R. Natarajan, A.~{Van der Ven}, and K.~Garikipati.
\newblock Machine learning materials physics: {I}ntegrable deep neural networks
  enable scale bridging by learning free energy functions.
\newblock {\em Computer Methods in Applied Mechanics and Engineering},
  353:201--216, 2019.

\bibitem{oommen2024rethinking}
V.~Oommen, K.~Shukla, S.~Desai, R.~Dingreville, and G.E. Karniadakis.
\newblock Rethinking materials simulations: Blending direct numerical
  simulations with neural operators.
\newblock {\em npj Computational Materials}, 10:145, 2024.

\bibitem{cohen2016group}
T.~Cohen and M.~Welling.
\newblock Group equivariant convolutional networks.
\newblock In {\em International conference on machine learning}, pages
  2990--2999. PMLR, 2016.

\bibitem{thomas2018tensor}
N.~Thomas, T.~Smidt, S.~Kearnes, L.~Yang, L.~Li, K.~Kohlhoff, and P.~Riley.
\newblock Tensor field networks: Rotation-and translation-equivariant neural
  networks for 3d point clouds.
\newblock {\em arXiv preprint arXiv:1802.08219}, 2018.

\bibitem{schutt2021equivariant}
K.~Sch{\"u}tt, O.~Unke, and M.~Gastegger.
\newblock Equivariant message passing for the prediction of tensorial
  properties and molecular spectra.
\newblock In {\em International conference on machine learning}, pages
  9377--9388. PMLR, 2021.

\bibitem{novick2008cahn}
A.~Novick-Cohen.
\newblock The {C}ahn--{H}illiard equation.
\newblock {\em Handbook of differential equations: evolutionary equations},
  4:201--228, 2008.

\bibitem{kovachki2023neural}
N.~Kovachki, Z.~Li, B.~Liu, K.~Azizzadenesheli, K.~Bhattacharya, A.~Stuart, and
  A.~Anandkumar.
\newblock Neural operator: Learning maps between function spaces with
  applications to pdes.
\newblock {\em Journal of Machine Learning Research}, 24:1--97, 2023.

\bibitem{evans2022partial}
L.C. Evans.
\newblock {\em Partial differential equations}, volume~19.
\newblock American mathematical society, 2022.

\bibitem{li2020fourier}
Z.~Li, N.~Kovachki, K.~Azizzadenesheli, B.~Liu, K.~Bhattacharya, A.~Stuart, and
  A.~Anandkumar.
\newblock Fourier neural operator for parametric partial differential
  equations.
\newblock {\em arXiv preprint arXiv:2010.08895}, 2020.

\bibitem{eyre1993systems}
D.J. Eyre.
\newblock {Systems of Cahn--Hilliard equations}.
\newblock {\em SIAM Journal on Applied Mathematics}, 53:1686--1712, 1993.

\bibitem{li2016multi}
Y.~Li, J.-I. Choi, and J.~Kim.
\newblock {Multi-component Cahn--Hilliard system with different boundary
  conditions in complex domains}.
\newblock {\em Journal of Computational Physics}, 323:1--16, 2016.

\bibitem{ten2024thermodynamically}
M.F.P. ten Eikelder, K.G. van~der Zee, and D.~Schillinger.
\newblock Thermodynamically consistent diffuse-interface mixture models of
  incompressible multicomponent fluids.
\newblock {\em Journal of Fluid Mechanics}, 990:A8, 2024.

\bibitem{garcke2003cahn}
H.~Garcke.
\newblock {On Cahn—Hilliard systems with elasticity}.
\newblock {\em Proceedings of the Royal Society of Edinburgh Section A:
  Mathematics}, 133:307--331, 2003.

\bibitem{oudich2025phase}
H.~Oudich, P.~Carrara, and L.~De~Lorenzis.
\newblock Phase-field modeling of elastic microphase separation.
\newblock {\em arXiv preprint arXiv:2505.01389}, 2025.

\bibitem{golovin1998convective}
A.A. Golovin, S.H. Davis, and A.A. Nepomnyashchy.
\newblock {A convective Cahn-Hilliard model for the formation of facets and
  corners in crystal growth}.
\newblock {\em Physica D: Nonlinear Phenomena}, 122:202--230, 1998.

\bibitem{balakrishna2018combining}
A.R. Balakrishna and W.C. Carter.
\newblock {Combining phase-field crystal methods with a Cahn-Hilliard model for
  binary alloys}.
\newblock {\em Physical Review E}, 97:043304, 2018.

\end{thebibliography}
\appendix
\section{Supplementary Results}

\subsection{Gradient-Enhanced Results}
\label{app:grad}

\begin{figure}[h!]
    \centering
    \includegraphics[width=\linewidth]{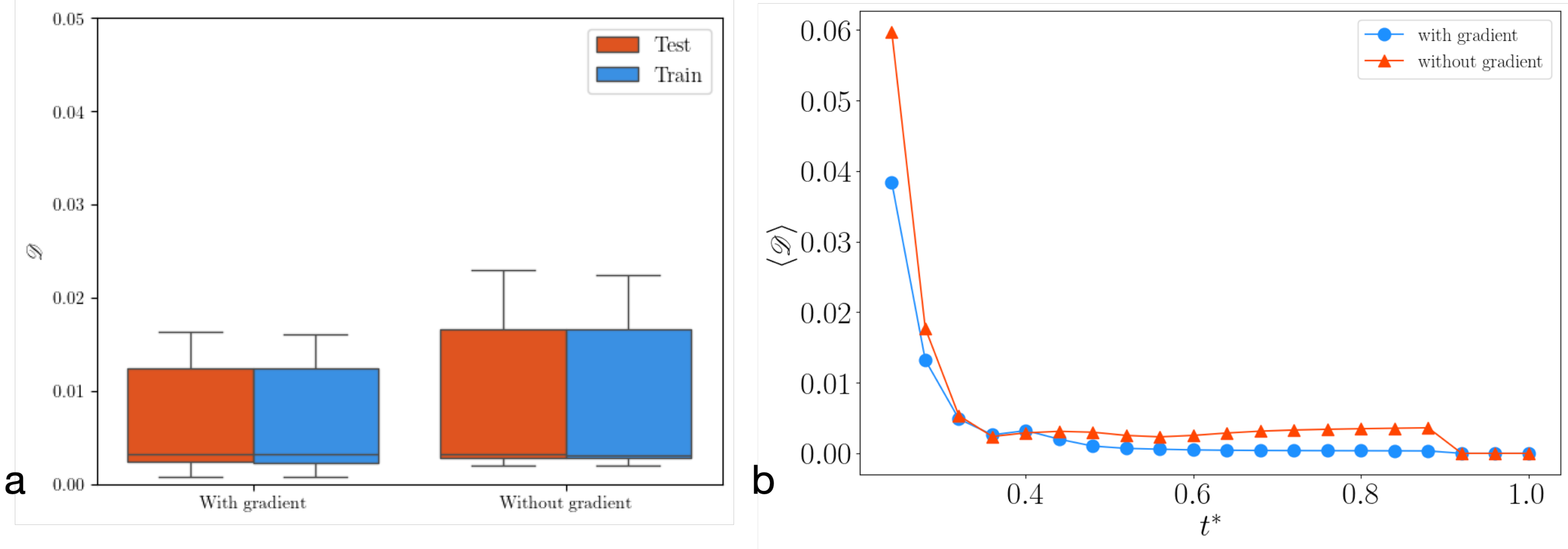}
    \caption{\textbf{a)} Box plots of the relative error $\mathscr{D}$ for training and test sets, with and without gradient loss. Gradient-informed training yields lower median errors and reduced variability. 
    \textbf{b)} Time evolution of the ensemble-averaged error $\mathscr{D}$ for models trained with $\mathcal{L}_{\text{norm}}$ (blue) and $\mathcal{L}_{\text{grad}}$ (orange). The gradient-enhanced model shows significant early-stage improvements.}
    \label{fig:gradient}
\end{figure}

By incorporating the spatial gradient term in~\cref{eq:loss}, we aim to guide the model toward better capturing interfacial dynamics. To assess its effectiveness, we compare the relative error $\mathscr{D}$ of models trained with and without the gradient term. Figure~\ref{fig:gradient} summarizes the performance comparison. The results demonstrate that incorporating spatial gradients leads to reduced variability and improved predictive stability, particularly during the early stages of microstructural evolution.


\subsection{Equivariance on FNO}\label{app:fno_equivariance}
\begin{figure}[h!]
    \centering
    \includegraphics[width=1\linewidth]{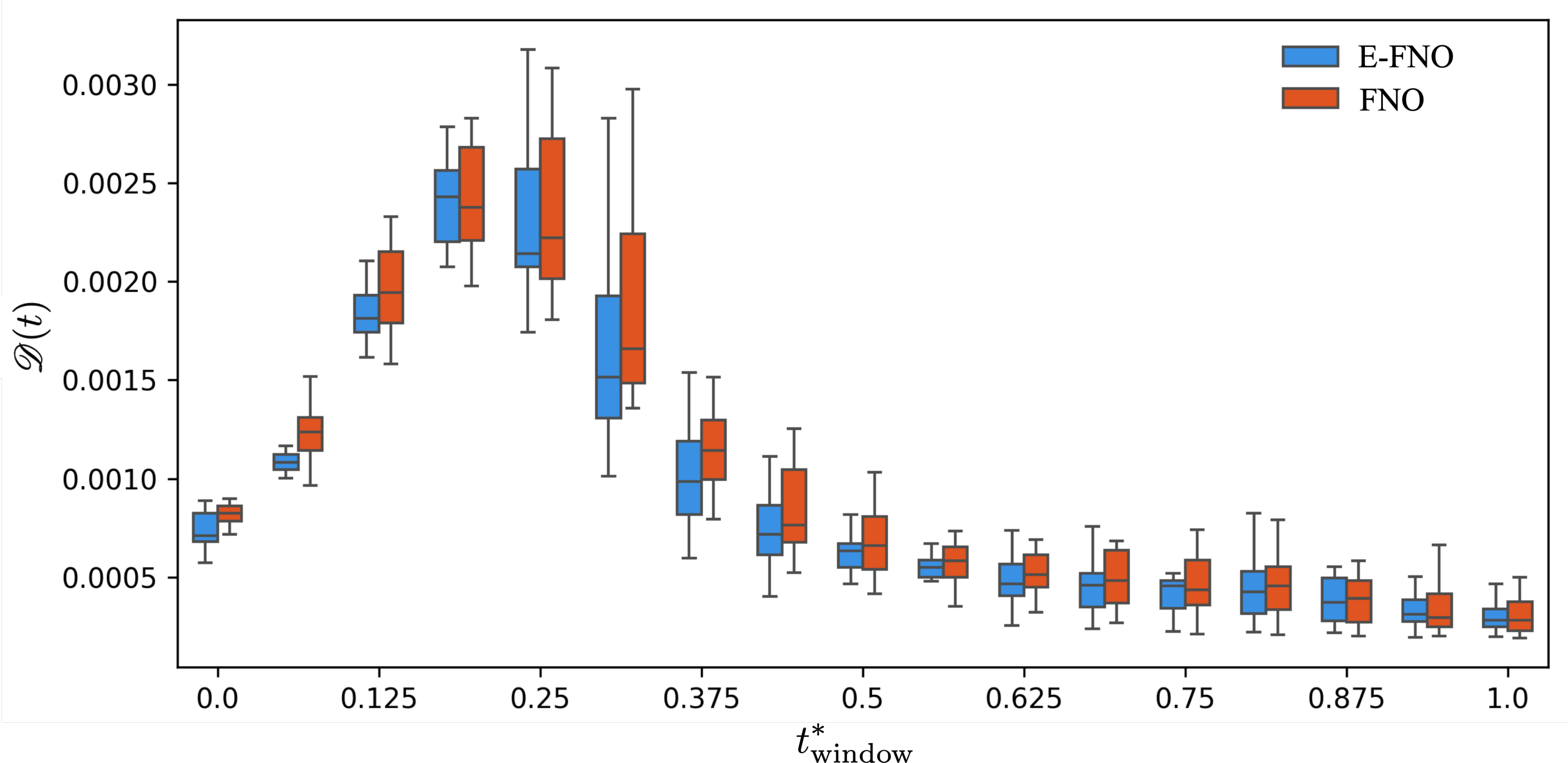}
    \caption{
    Distribution of the relative $L_2$ loss $\mathscr{D}(t)$ over successive time windows for the Fourier Neural Operator (FNO) trained with (blue) and without (orange) the $D_4$-equivariance loss term. Each box plot shows the error distribution within a time window across all spatial points, with whiskers representing 1.5$\times$ the interquartile range and circles indicating outliers. Incorporating equivariance consistently reduces the median error and variability during the early and most dynamic stages of evolution (time windows 1–7), while performance differences diminish as the system approaches equilibrium.
    }
    \label{fig:fno_eq_loss_comparison}
\end{figure}
Figure~\ref{fig:fno_eq_loss_comparison} presents a comparison of the Fourier Neural Operator trained with and without the $D_4$-equivariance described in Section~\ref{sec:E-UNO}. The plotted metric, $\mathscr{D}(t)$, represents the relative $L_2$ error between predicted and reference fields within each time window. In the early to mid stages of the simulation (time windows 1–7), where morphological changes are rapid and interface dynamics are most pronounced, the equivariant FNO exhibits lower median errors and reduced spread compared to the baseline. These improvements highlight the benefit of symmetry constraints in guiding the network toward physically consistent solutions. In the later stages, where the microstructure coarsens and the system approaches equilibrium, the error levels of both models converge, indicating that the impact of equivariance is most significant during high-activity phases of the evolution.

\subsection{Systematic comparison between different Neural operators}\label{sec:system}

The inference times on a single 80G-VRAM A100 NVIDIA GPU of E-UNO (UNO) and E-FNO (FNO) are shown in Table \ref{tab:inference_times}, which demonstrates that UNO has a slight advantage over FNO in terms of inference speed. The model hyperparameters are listed in Table \ref{tab:uno_fno_hparams}. 

\centering
\begin{table}[h!]
\centering
\begin{tabular}{|c|c|}
\hline
\textbf{E-UNO and UNO Inference Time} (seconds) & \textbf{E-FNO and FNO Inference Time} (seconds) \\
\hline
$0.030 \pm 0.002$ & $0.034 \pm 0.001$ \\
\hline
\end{tabular}
\caption{Comparison of inference times between UNO and FNO. The values are the average over the 
inference time of 1000 cases with $\pm$ value indicates the standard deviation}
\label{tab:inference_times}
\end{table}

\begin{table}[h!]
\centering
\begin{tabular}{|l|p{7cm}|p{5cm}|}
\hline
\textbf{Hyperparameter} & \textbf{E-UNO \& UNO} & \textbf{E-FNO \& FNO} \\
\hline
Initial Learning Rate ($lr_{init}$) & $5 \times 10^{-4}$ & $5 \times 10^{-4}$ \\
\hline
Final Learning Rate ($lr_{final}$) & $1 \times 10^{-5}$ & $1 \times 10^{-5}$ \\
\hline
Output Channels & [32, 64, 64, 128, 64, 64, 32] & -- \\
\hline
Modes & [[32, 32], [16, 16], [8, 8], [4, 4], [8, 8], [16, 16], [32, 32]] & (16, 16) \\
\hline
Scalings & [[1.0, 1.0], [0.5, 0.5], [0.5, 0.5], [1, 1], [2, 2], [2, 2], [1, 1]] & -- \\
\hline
Model total parameters & 6,376,406 & 6,288,090 \\
\hline
\end{tabular}
\caption{Summary of hyperparameters for UNO and FNO models. The models were trained using a cosine 
annealing learning rate schedule over 200 epochs. The parameters not listed here are the default values in \texttt{neuraloperator} python package.}
\label{tab:uno_fno_hparams}
\end{table}

\end{document}